\documentclass{article}

\usepackage{arxiv}

\usepackage[utf8]{inputenc} 
\usepackage[T1]{fontenc}    
\usepackage{hyperref}       
\usepackage{url}            
\usepackage{booktabs}       
\usepackage{amsfonts}       
\usepackage{nicefrac}       
\usepackage{microtype}      
\usepackage{graphicx}
\usepackage{listings}
\usepackage{svg}
\usepackage{xcolor}
\usepackage{amsmath}
\usepackage{natbib}
\graphicspath{ {./images/} }

\title{From Judgments to Issues: Structured Extraction of Legal Reasoning with Citation-Hallucination Control}

\author{
 Giovanni Piccioli \\
  Quantitative and Digital Law Laboratory\\
  King's College London\\
  Strand, London WC2R 2LS \\
  \texttt{giovannipiccioli@gmail.com} \\
   \And
 Alessia Fidelangeli \\
  CIRSFID - Alma AI, Faculty of Law\\
  University of Bologna\\
  Bologna, Via Zamboni 27/29 \\
  \And
 Piera Santin \\
  Robert Schuman Centre\\
  European University Institute\\
  Fiesole, Badia Fiesolana - Via dei Roccettini 9 \\
  \And
 Pierpaolo Vivo \\
  Quantitative and Digital Law Laboratory\\
  King's College London\\
  Strand, London WC2R 2LS \\
}

\begin{document}
\maketitle
\begin{abstract}
We present an automated pipeline that decomposes Italian tax-court judgments into individual legal issues and extracts, for each issue, a structured XML representation grounded in the IRAC framework and the legal syllogism. The pipeline targets a corpus of approximately $330{,}000$ first- and second-instance decisions of the Italian tax courts and is built around a capable yet cost-efficient general-purpose model (DeepSeek V3), a choice driven by the need to process several hundred thousand documents at a sustainable cost. To address the well-documented unreliability of large language models on legal citations, we couple the extraction step with an automatic hallucination-detection filter that compares the references produced by the model with those identified in the judgment text by a dedicated parser (Linkoln), normalised to standard identifiers (URN-NIR, ECLI, CELEX). We validate the pipeline on $50$ judgments annotated by two PhDs in tax law, computing inter-annotator agreement and LLM-vs-expert agreement on both issue extraction and legal citations, together with a stand-alone evaluation of the hallucination filter. To the best of our knowledge, this is the first issue-level, expert-validated structured extraction  pipeline with hallucination control for Italian tax-court decisions, and it provides a concrete starting point for downstream applications such as issue-level retrieval, citation-network analysis, and the construction of large-scale datasets of legal reasoning.
\end{abstract}


\section{Introduction}
\label{sec:introduction}

Modern judicial systems generate court decisions at a scale that has long outpaced the capacity of legal professionals to read them individually. In Italy, the database of first- and second-instance tax-court decisions alone contains almost a million judgments, and grows by the hundreds of thousands every year. The practical value of this volume lies almost entirely in secondary use: precedent search by practitioners, citation-network analysis, evaluation of judicial behaviour, and the construction of large datasets for training and benchmarking legal NLP systems. Each of these uses, however, requires a representation of the decisions that is more compact and structured, than the original PDFs the courts produce. The central question this paper addresses is how to obtain such a representation automatically, at scale, and with quantified reliability.

Italian tax-court judgments are a particularly informative testbed for this question. On the one hand, the domain stresses any extraction pipeline that claims to be general. Tax disputes range from a few tens of euros to tens of millions, and routinely draw on civil, commercial, criminal, procedural, and European Union law in resolving a single case. The first two instances are decided by panels that include non-career honorary judges, whose drafting style is markedly less standardized than that of higher courts: even the names and ordering of the structural sections of a judgment vary from one decision to the next.
On the other hand, the domain also makes the problem tractable: an official open database is available, the volume is large enough that statistical evaluation is meaningful, and the parties are always the same pair (taxpayer and tax authority). A pipeline that performs well here is therefore informative both as a methodological contribution and as a usable tool for a jurisdiction whose case law has, so far, received little attention in the legal NLP literature.

The central methodological choice of this work is to take the \emph{legal issue} (\emph{questione giuridica}), rather than the judgment, as the unit of representation. This choice is consistent with long-standing analyses of judicial justification, which view adjudication as the resolution of one or more legal questions through  reasoning grounded in facts and legal norms \citep{maccormick1994legal, bench2024computational}. A single judgment typically resolves several issues that have little in common beyond belonging to the same case: a court may first decide on the admissibility of an appeal, then on the merits, and finally on the allocation of costs. For practically all secondary uses we have in mind, the issue is the natural unit of analysis: a precedent search is a search for cases that have decided a specific question; a citation graph is informative to the extent that its edges reflect references actually used to decide a point of dispute, rather than incidental mentions; a dataset of legal reasoning is useful to the extent that its examples isolate a single argumentative chain. We therefore design our pipeline to decompose each judgment into a set of autonomous issues so as to operate on these issues independently.

The structure we extract for each issue is inspired by two widely used analytical frameworks for describing judicial reasoning: IRAC (Issue, Rule, Application, Conclusion) and the legal syllogism. These models impose enough structure to make the representation machine-readable and to anchor it in a recognized theoretical tradition, while remaining coarse-grained enough to be extracted reliably by a language model. Concretely, each issue is encoded as an XML record with a \textit{Whether}-clause statement of the question, a list of factual premises, a list of legal references with their citation reasons, the judge's reasoning, the outcome, and an abstractive summary. The schema deliberately sits at a level of granularity that users can interpret easily, and that the language model can populate consistently across the heterogeneous styles of the corpus. The trade-off between structural richness and reliable extractability is itself a design choice as discussed in detail in Section~\ref{sec:balancing}.

Two practical constraints shape the rest of the pipeline. The first is cost. Extraction at the scale of the full database requires a model whose per-document price is compatible with processing several hundred thousand judgments; this motivates our use of DeepSeek V3 which resulted in a cost of roughly \$0.0035 per judgment. In comparison, the rates of top-range proprietary models can be twenty times higher. The second is reliability. The LLM should be powerful enough to correctly segment the judgment into issues and fill the corresponding XML fields in a way that a tax law expert would find acceptable. In addition, the model's hallucination rate should be as low as possible. This is a known failure mode of LLMs, consisting in the fabrication of plausible-looking but non-existent citations to legislation and case law ~\citep{dahl2024large,han2026legal}.

Because cost-efficiency and reliability represent competing priorities, we conducted an informal comparative assessment of multiple models and selected the most cost-effective model that delivered results acceptable to legal experts; see Appendix~\ref{app:model_selection}.
To mitigate the risk of hallucinations, we add an explicit verification layer after the extraction. We parse the legal references in both the judgment text and the LLM output with the Linkoln library~\citep{linkoln}, normalise them to standard identifiers (URN-NIR, ECLI, CELEX), and remove from the output any reference that cannot be located in the source. This filter is conservative by design: it preserves the citations on which the pipeline is reliable and removes those it is not.

We evaluate the resulting pipeline on a manually annotated subsample of $50$ judgments. Two experts (PhDs in tax law) doubly annotated $20$ of these under double-blind conditions and annotated the remaining $30$ in disjoint halves, following guidelines they had jointly developed on a separate set of judgments. This protocol lets us report LLM-vs-expert agreement on issue extraction, on legal references extraction, and on a set of qualitative scores for the free-text fields, and also inter-annotator agreement on the same metrics. Section~\ref{sec:results} reports the results in detail; we additionally evaluate the hallucination filter as a stand-alone classifier and analyze the residual hallucinations by citation type.

\subsection{Contributions}
\label{sec:contribution}
The contribution of this paper is threefold.

First, we propose a structured XML schema for representing court judgments that is explicitly grounded in IRAC and in the legal syllogism, takes the legal issue as its unit, and is simultaneously useful for human consultation, machine-readable for downstream tasks, and reliably instantiated by a mid-range LLM.

Second, we develop an extraction pipeline built around a cost-efficient general-purpose model (DeepSeek V3) and combine it with an automatic hallucination-detection step on legal references. By normalizing both the judgment text and the LLM output through the Linkoln library~\citep{linkoln} and comparing the two sets, we remove citations that the model has fabricated while preserving the rest. The pipeline is run and refined on a set of about $330{,}000$ processed (of $401{,}349$ retrieved), Italian tax court judgments.

Third, we evaluate the pipeline on a corpus of $50$ Italian tax-court judgments annotated by two legal experts, reporting inter-annotator agreement and LLM-vs-expert agreement on issue extraction, on legal citations, and on qualitative dimensions of the free-text fields, together with a stand-alone analysis of the hallucination filter. 

The rest of the paper is organised as follows. Section~\ref{sec:related_work} reviews related work on structured extraction from court decisions, IRAC- and syllogism-based pipelines, citation handling, and LLM-based legal annotation. Section~\ref{sec:schema} presents the issue-based XML schema and the design requirements that shape it. Section~\ref{sec:pipeline} describes the corpus and the extraction pipeline, including the hallucination-detection step. Section~\ref{sec:validation_method} sets out the validation protocol, and Section~\ref{sec:results} reports the evaluation results. Section~\ref{sec:conclusion} concludes with a discussion of limitations and directions for future work.

\section{Related Work}
\label{sec:related_work}

Our work sits at the intersection of several lines of research in legal NLP: (i) formal models of judicial reasoning, (ii) structured extraction from court decisions, (iii) legal citation extraction, normalization, and hallucination control, and (iv) the use of LLMs for extraction and annotation on legal corpora.

\paragraph{IRAC, the legal syllogism, and structured judicial reasoning.}
A first body of work in AI and Law seeks to represent legal reasoning in structured forms that support computational analysis, legal knowledge extraction, and decision support \citep{ashley2017artificial}.
Within this tradition, several recent works formalize judicial reasoning along the lines of the IRAC framework or of the legal syllogism and exploit these structures in LLM-based pipelines \citep{bench2020explaining}. \citet{jiang2023legal} introduce \emph{legal syllogism prompting}, instructing LLMs to articulate major premise, minor premise, and conclusion for judgment prediction; \citet{yu2022legal} similarly show that IRAC-style prompts outperform generic chain-of-thought on the legal entailment task in the COLIEE competition \citep{rabelo2022overview}. \citet{kang2025automating} couple an IRAC benchmark of Malaysian contract-law scenarios with a semi-structured knowledge base, showing substantial gains on issue identification and rule retrieval. \citet{holzenberger2023connecting} extend this approach by treating statutory reasoning as a pipeline in which legal text is first converted into structured representations through information extraction, and then processed by a symbolic Prolog-based reasoner, showing that reasoning accuracy depends on the quality of the extracted structure. While these works use IRAC or the syllogism to guide prompting or give LLMs a framework for legal reasoning, we take inspiration from them in designing the extraction schema \citep{ruf2024aristotle}.

\paragraph{Structured extraction and decomposition of judgments.}
A closely related strand uses LLMs to extract structured representations of decisions for secondary use. \citet{grundler2025automated} extract Judicial Interpretative Formulas from CJEU decisions on VAT, combining LLM-based annotation with fine-tuned BERT and expert validation; their tax-law focus and expert-validated setup parallel ours. \citet{adhikary2024case} define fine-grained attributes for Indian criminal cases and extract them with few-shot LLM prompting, demonstrating that attribute-level representations help with judgment and statute prediction. \citet{costa2024automated} produce structured annotations of Brazilian judicial decisions tagged with \emph{ratio decidendi} and outcome; \citet{westermann2025automated} use LLMs to map court decisions onto a "rule tree" (a logical breakdown of the conditions a claim must satisfy) identifying which conditions the judge found met and explaining the reasoning; \citet{janatian2023text} convert legislation into structured if-then rules that computer systems can use to answer legal questions. \citet{belfathi2023harnessing} and \citet{lombardi2023legal} use LLMs to segment and label judgments into logical parts. The annotation of arguments in judgments and subsequent use for NLP tasks is explored in  \citet{habernal2024mining} and  \citet{santin2023argumentation}. Finally, the Openjustice initiative \citep{dahan2023openjustice, bhambhoria2024evaluating} aims to build a structured dataset or legal reasoning based on crowdsourcing reasoning flows composed of facts, logic conditions, and outcomes.
Compared with this body of work, our contribution differs in (i) anchoring the extraction template in IRAC/syllogism rather than in a flat attribute list, (ii) decomposing each judgment into autonomous \emph{issues}, and (iii) integrating an explicit hallucination check on legal references.

\paragraph{Legal holdings and summarization, with an Italian focus.}
The extraction of holdings, \textit{massime}, and \textit{Leitsätze} and structured summaries is closely related to ours. \citet{licari2023legal} conducted experiments aiming to extract \textit{massime} from Italian judgments using Italian-LEGAL-BERT. \citet{zin2025court} study LLMs for generating German \textit{Leitsätze} and \citet{arvin2025identifying} systematically evaluate LLMs of different sizes for holding identification. \citet{xu2023question} propose evaluating legal summaries by whether they address every legal issue raised in the underlying judgment. For Italian,\citet{benedetto2023benchmarking,benedetto2025legitbart} benchmark and develop BART-based summarization models adapted to Italian legal documents. Most directly comparable, \citet{pont2023legal} apply GPT-4 to Italian tax-court judgments within the PRODIGIT project for summarization, identification of legal issues, and keyword extraction, with extensive evaluation by tax judges and lawyers.\citet{garzo2025does} evaluate GPT-4o on Italian vehicular homicide rulings, assessing both conceptual understanding and the ability to extract the \emph{ratio decidendi}. We extend these contributions by moving from free-form abstractive summaries to a fully structured XML schema modelled on IRAC, decomposing each judgment into individual issues with their own fields and embedding citation-level hallucination control.

\paragraph{Citation extraction, normalization, and citation networks.}
Reliable handling of legal references is a fundamental element of our pipeline. The Linkoln library~\citep{linkoln} parses legal citations in Italian and resolves them to standard identifiers (urn:nir, CELEX, ECLI); the URN-NIR scheme for Italian legal acts is described in \citet{francesconi2010urn} and \cite{digitpa2012linee}. We rely on these tools to standardize citations in both the judgment text and the LLM output, and to flag references that do not match as candidate hallucinations. Several works focused on building legal citation networks, see for example \cite{coupette2021measuring}, \cite{ovadek2021analysing}. The construction of citation networks from judgments has been recently revisited by \citet{mumford2025context}, who annotate each citation \emph{instance} with its context; our \verb|<citation_reason>| field is aligned with this idea.

\section{An Issue-Based Schema for Tax-Court Judgments}
\label{sec:schema}

\subsection{The XML format}
\label{sec:xml_format}
In this section we motivate and describe the different fields we extract from each judgment. Our work in fact starts from the idea of transforming a judgment into another structured document. The choice of this structure is crucial, as it determines the possible uses of the output format. We selected a possible format based on the following criteria:
\begin{enumerate}
    \item Usefulness for legal professionals. A professional conducting a precedent search looking at the extracted fields should more easily grasp the essential information about the case, compared to having to read the original document.
    \item Possibility of extracting the information automatically using LLMs. In other words, the format should be simple enough to be reliably filled by an LLM, when the text of a judgment is provided.
    \item Possibility to use the extracted format for computational tasks. In particular, we value the machine readability of extracted fields.
    \item Generality. The format should be general enough to apply to several types of judgments.
\end{enumerate}
In essence, our goal is to develop a structured format that can be extracted automatically from the text of judgments and that is useful for secondary use, both for human use and for computational analysis. In practice, there can be tension between some of the requirements, for example machine and human readability. 
To obtain the final structure, we iteratively refined an extraction prompt through several rounds of validation.
To design the structure, we were also inspired by the IRAC approach and by the legal syllogism. These frameworks give a stylized view of the judge's reasoning process. In listing \ref{lst:prompt} we report an English translation of the final prompt we used for the extraction. The original Italian prompts are reported in Appendix~\ref{app:prompts_ita}.
\begin{lstlisting}[caption={Prompt (translated in English) used for XML extraction.}, label={lst:prompt}]
Inside the curly braces there is an Italian tax court decision.
Your task is to identify and extract the main legal issues (typically one or two) that the judge addresses and resolves in order to decide the dispute. Consider only the issues on which the judge rules explicitly and with reasoning.

For each issue, perform the following:
- Formulate the legal issue clearly, as a self-contained and autonomous question. The formulation must follow the structure Whether + clause and must be understandable on its own, without referring to the full decision or other issues.
- Indicate the outcome of the issue, i.e., the answer to the legal question.
- Break down the judge's reasoning into the following parts:
  -- <factual_premises>: list the facts, actions, events, or data of the case that are relevant to the issue.
  -- <legal_references>: list ALL legal references (legislation, case law, general principles, administrative practice) relevant to the issue, as they appear in the decision. Order them by decreasing importance.
  -- <citation_reason>: for the legal references discussed more in detail, explain why the judge cites them. Include only references that are analyzed in depth.
  -- <judge_reasoning>: describe in free text how the judge applied the legal references to the concrete case, leading to the decision. Do not repeat the factual premises.
- Write an abstractive summary of the issue (100-150 words) including: facts, law, reasoning, and conclusion. The summary must be self-contained and understandable without referring to the full decision or other issues.

General requirements:
- Each issue must be autonomous, self-contained, and formulated in precise and detailed legal language.
- The facts must be described with sufficient precision to allow another judge to decide a similar issue consistently.

Required output format: XML only, as in the following schema:
<issues>
  <issue title="issue title">
    <text>text of the legal issue (~30 words)</text>
    <issue_outcome in_favor_of="tax authority|taxpayer|other">outcome text</issue_outcome>
    <factual_premises>
      <item>factual premise 1</item>
      <item>factual premise 2</item>
    </factual_premises>
    <legal_references>
      <item id="D1" type="caselaw" ref="Supreme Court no. nnnn/yyyy"/>
      <item id="D2" type="leg" ref="art. x law no. nnn/yyyy"/>
    </legal_references>
    <citation_reason>
      <item ref_id="D1">reason for citing legal reference</item>
      <item ref_id="D2">reason for citing legal reference</item>
    </citation_reason>
    <judge_reasoning>reasoning text (~100 words)</judge_reasoning>
    <summary>summary text (~100 words)</summary>
  </issue>
</issues>
In the "ref" field of legal references, always indicate the exact citation of the act. Use the "type" field as follows:
- type="caselaw" for case law. If multiple decisions are cited consecutively, group them into a single item;
- type="leg" for legislation;
- type="princ" for general legal principles (in this case, "ref" contains only the name of the principle);
- type="admin_pract" for administrative practice.

Output ONLY the XML block and nothing else.
\end{lstlisting}

The prompt instructs the LLM to extract the main issues contained in the judgment. In addition, the prompt also instructs the model to keep issues autonomous. In other words each issue should be self-standing and understandable on its own. Our format therefore aims to decompose a judgment into a set of independent issues and to treat each separately. This is justified by the fact that judgments are often heterogeneous entities. Within one judgment, the judge might first decide on the admissibility of the appeal; then once this is declared admissible it goes on to decide on the merits of the case. These would correspond to two separate issues that indeed have little reason to be grouped together for secondary use. The XML format still preserves the unity of the judgment by grouping all issues belonging to the same judgment in the \verb|<issues></issues>| tags.

For each issue, the LLM is instructed to extract several fields:
\begin{itemize}
    \item \verb|<text>|, the question of the issue, formulated in a standardized way 'Whether + clause'. 
    \item \verb|<issue_outcome>| the outcome of the issue, i.e., the judge's decision on the matter. The attribute \verb|in_favor_of| taking values in (\verb|taxpayer, tax authority, other|) provides a simple classification of the outcome.
    \item \verb|<factual_premises>| a list of relevant facts to understand the case
    \item \verb|<legal_references>|, and  \verb|<citation_reason>| a list of legal citations, each optionally accompanied by a short sentence describing why it was cited. Each reference is assigned a type among caselaw, legislation, principle, administrative practice.
    \item \verb|<judge_reasoning>| a textual field focused on reporting the application of the law to the particular case.
    \item \verb|<summary>| a textual summary of the overall issue, with potential overlaps with other fields. 
\end{itemize}
Listing~\ref{lst:snail_farmer} gives a compact view of an issue extracted from judgment n.~318/2024 of the Corte di Giustizia Tributaria di primo grado di Teramo; the full extraction, with all fields reported verbatim, is reproduced in Appendix~\ref{app:worked_example}.

\begin{lstlisting}[caption={Compact view of an extracted XML issue (full version in Appendix~\ref{app:worked_example}).}, label={lst:snail_farmer}]
<issues>
  <issue title="VAT deductibility for an agricultural business not yet productive" id="Q1">
    <text>Whether the purchase of a capital asset (agricultural tractor) by a newly established
          agricultural business, not yet productive, is connected to the entrepreneurial activity
          and therefore deductible [...].</text>
    <issue_outcome in_favor_of="taxpayer">VAT is deductible because the purchase is related to
          the entrepreneurial activity, even in the absence of immediate taxable operations.</issue_outcome>
    <factual_premises>
      <item>The appellant began the snail-farming activity in 2020, an activity that requires
            time to become productive.</item>
      <item>In 2022, the appellant purchased an agricultural tractor for land preparation and
            for procuring plant-based feed necessary for the farming activity.</item>
      <item>[... 2 further factual premises ...]</item>
    </factual_premises>
    <legal_references>
      <item id="D1" type="caselaw" ref="Cassazione n. 26689/2022; n. 7440/2021; n.15570/2023"/>
      <item id="D4" type="leg" ref="Articles 4 and 5 DPR 633/72"/>
      <item>[... 3 further legal references ...]</item>
    </legal_references>
    <citation_reason>
      <item ref_id="D1">National case law states that the right to deduct VAT is linked to the
            relevance of the asset to the entrepreneurial activity, even in the absence of
            immediate taxable operations, provided the asset is functional to the organisation
            of the business.</item>
      <item>[... 2 further citation reasons ...]</item>
    </citation_reason>
    <judge_reasoning>The judge considers that the purchase of the tractor is connected to the
          taxpayer's agricultural activity, despite the absence of immediate taxable operations
          [...]. Therefore, the denial is annulled and the VAT refund is granted.</judge_reasoning>
    <summary>The taxpayer, who operates a snail-farming business, purchased a tractor in 2022
          to prepare the land, despite production not yet having begun. [...] The ruling
          annulled the denial and recognised the deductibility.</summary>
  </issue>
</issues>
\end{lstlisting}

The example illustrates how each field of the schema is instantiated on a real decision. A single \verb|<issue>| is extracted in this case.

\subsection{IRAC and the legal syllogism}
\label{sec:irac_syllogism}
We now provide a comparison between our structured format and the IRAC and legal syllogism frameworks, to which our format is inspired. IRAC is a structured legal reasoning framework that stands for 'Issue' (the legal question at stake), 'Rule' (the applicable law or precedent), 'Application' (how the rule applies to the facts), and 'Conclusion' (the decision). The parallel of our format with IRAC starts with the idea of extracting issues. Some of the different XML fields resemble the IRAC elements as follows. 
\begin{itemize}
    \item Issue: the \verb|<text>| field states the issue. In addition \verb|<factual_premises>| provides some additional context.
    \item Rule: the references in \verb|<legal_references>| contain the relevant rules.
    \item Application: the field \verb|<judge_reasoning>| details how the legal references are applied. In addition, the items in \verb|<citation_reason>| provide some additional information on the reasons why individual references were quoted.
    \item Conclusion: \verb|<issue_outcome>| contains the decision on the issue. The information is partially present also in the \verb|<judge_reasoning>| field.
\end{itemize}

However, there are differences between IRAC and our format. First, the \verb|<legal_references>| field disaggregates the 'Rule' component by assigning each citation a type (case law, legislation, principle, administrative practice) and an optional \verb|<citation_reason>|, making explicit the distinct role each source plays in the reasoning. Second, IRAC does not distinguish between the facts of the case and the legal issue, while our format separates \verb|<factual_premises>| from \verb|<text>|. Finally, the field \verb|<summary>| has no direct counterpart in IRAC.

Similarly, we can provide a parallel with the legal syllogism. A legal syllogism on a given issue is composed of the following parts:
\begin{enumerate}
    \item Major premise (rule): parallel with \verb|<legal_references>|.
    \item Minor premise (facts): parallel with \verb|<factual_premises>|.
    \item Conclusion (judgment): similar to \verb|<issue_outcome>|.
\end{enumerate}
Although legal syllogism is a useful reference, the idea that judges strictly follow syllogistic reasoning is debated in the literature on legal theory. Moreover, it is far from evident in unstructured texts, such as tax court decisions, where the reasoning is often implicit, fragmented, or intertwined with procedural matters. Our format is therefore \textit{inspired by} the syllogism: in designing the fields we had syllogism in mind, but the structure was refined in several respects. First, the fields are more granular than the syllogistic parts. Second, and most importantly, a key component of our format is the \verb|<judge_reasoning>| field, which captures forms of reasoning that go beyond the syllogistic model, such as reasoning by analogy, balancing of competing principles, or teleological interpretation.  Unlike the major and minor premises of a syllogism, which are based on a fixed logical relationship, both \verb|<legal_references>| and \verb|<factual_premises>| can contain multiple elements on a single issue, without a one-to-one correspondence between them. 
This is linked to our format adopting a \textit{single-step} model per legal issue. In principle, the reasoning leading to the decision of an issue could be decomposed into a \textit{chain of syllogisms}, where each step uses the conclusions of the previous ones as a minor premise, and the conclusion of the last step coincides with the decision on the issue. Such a multi-step decomposition would yield a fine-grained picture of the reasoning. However, this structure comes at the price of increased ambiguity in extraction and lower reproducibility. Hence, in our format, each legal issue is associated to a single structured representation, which captures the relevant rules, facts, reasoning, and outcome without decomposing the inferential path into intermediate steps. This design choice prioritises consistency and extractability.

\subsection{Design pressures and potential uses}
\label{sec:design_pressures}
The schema described above is shaped by the requirements imposed by four anticipated uses of the extracted format. We discuss each in turn, in each case naming the design choice that follows from the corresponding requirement.

\paragraph{Human consultation.}
In the context of precedent research, as an alternative to reading the text of the judgment, the user may find it easier to grasp the central points of the judgment by looking at a graphical interface displaying the XML fields of one of its issues. The decomposition into issues also means that the user can engage directly with the issue they are interested in, without having to deal with other potentially irrelevant parts of the judgment.
The format would then play a role similar to that of headnotes, or \textit{massime} in the case of Italian judgments. However, differently from \textit{massime}, which select issues based on their novelty and relevance to the uniform interpretation of the law (\textit{funzione nomofilattica}), our pipeline aims to extract all issues addressed in a judgment, including repetitive or well-settled ones. In practice, however, we instruct the LLM to focus on the most relevant issues, as extracting the full set of legal issues from every decision can be beyond its reliable capabilities.

\paragraph{Issue-level retrieval.}
The second use is automated retrieval over the corpus. Existing legal information systems combine full-text and semantic methods, both of which suffer when the unit of retrieval is the full judgment. Judgments are long and contain substantial amounts of structurally necessary but retrieval-irrelevant text (standard procedural formulas, section headings, recurring boilerplate) which introduces noise into the document representation and degrades the precision of full-text search. Semantic retrieval, in turn, requires documents to be segmented into (often overlapping) chunks and indexed at the chunk level, which increases index size and redundancy and may obscure long-range dependencies when relevant information is distributed across chunks. A representation that supports retrieval well must therefore (i) isolate single legal questions and (ii) produce self-contained units of appropriate length. The decomposition into issues and the use of an abstractive \verb|<summary>| for each issue address both requirements: each extracted issue is a self-contained unit, several hundred tokens long, that strips boilerplate and structurally irrelevant content while preserving the legal context.

\paragraph{Citation networks.}
The legal references attached to each extracted issue can be used to construct a citation network between judgments. To this end, a parser is first employed to standardize citation formats. The resulting network can be further enriched by associating each edge with the textual attributes from  \verb|<citation_reason>|, capturing the reasons for which a given document is cited. When a document is cited multiple times across judgments, the aggregation of these attributes over incoming edges provides a useful indication of its practical relevance, highlighting the most common purposes for which the document is invoked. The construction of such a graph also allows to implement advanced retrieval algorithms such as PageRank \citep{brin1998anatomy}. 
A natural objection is that such a network could be constructed directly from judgments without relying on issue-level extraction. The key difference lies in the set of citations considered. A network built from full judgments includes all references appearing in the decision, many of which are not central to the court's reasoning. In Italian tax law judgments, for example, parties often refer to case law or statutory provisions in the section reporting their submissions (\textit{richieste delle parti}), which are not necessarily taken up in the court's reasoning. More generally, decisions frequently include references to standard or procedural provisions (e.g. jurisdictional rules or general principles) that serve mainly a formal function and add little to the resolution of the dispute. By contrast, when citations are extracted in association with individual issues, the resulting network is still defined at the judgment level, but its edges are based only on references that are relevant to the resolution of those issues. This selective inclusion improves the signal-to-noise ratio of the network, yielding a representation that more accurately reflects the relationships between decisions. Appendix~\ref{app:linkoln_baseline} assesses this argument empirically on the validation set, comparing the citations parsed by Linkoln from the full text and from the reasoning section with the expert-validated issue-level references.

\paragraph{Legal reasoning dataset.}
The proposed format can be seen as a simplified and operational model of legal reasoning. While numerous theoretical models of legal reasoning have been proposed in the literature, there is a lack of large-scale, structured datasets that instantiate these models in practice. By extracting issue-level representations inspired by IRAC and syllogistic structures, our approach provides a first step toward the creation of such datasets. A corpus of extracted issues could support the training and evaluation of machine learning models for tasks involving legal reasoning, offering a connection between abstract theoretical frameworks and data-driven methods.

\subsection{Balancing Structure, Expressiveness, and Extractability}
\label{sec:balancing}
A central design challenge in defining the proposed XML schema concerns the trade-off between structural rigidity and expressive adequacy. On the one hand, a relatively rigid format is desirable to ensure machine readability and facilitate computational use: for instance, modeling each issue as a single-step judicial syllogism enables consistent extraction of premises, legal references, and outcomes, at the cost of abstracting away from more complex reasoning patterns (e.g. chains of interdependent syllogisms). On the other hand, secondary use by human users often benefits from more flexible, narrative representations (such as the summary field) which can capture nuances that are difficult to encode in strictly structured components. A further tension arises from the heterogeneity of the underlying corpus: Italian tax law judgments exhibit substantial variation in structure and style. An overly restrictive schema risks forcing the extraction of elements that are not explicitly present in the text, thereby increasing the likelihood of hallucinations or unstable outputs.  Similarly, an overly rich or complex schema may introduce ambiguity, leading to inconsistent representations even across expert annotators. These considerations are compounded by the practical limitations of LLM-based extraction, as increasingly complex templates place greater demands on the model's ability to produce reliable outputs. The proposed format reflects the compromise objective to provide enough structure for computational use while maintaining flexibility to accommodate variability in the source material and to support meaningful human interpretation.

\section{Extraction Pipeline}
\label{sec:pipeline}

\subsection{Dataset}
\label{sec:dataset}
The corpus used in this work consists of first- and second-instance Italian tax-court judgments retrieved from the official database of tax-court decisions \cite{banca_dati_giurisprudenza_tributaria}. At the time of extraction we retrieved almost all the available decisions published between 2021 and 2023, amounting to  $401{,}349$ documents.

Judgments are distributed by the database in PDF format. We converted each document to plain text using the PyMuPDF library \cite{pymupdf}. The resulting texts are heterogeneous in layout and typographic conventions, reflecting the absence of a standardized drafting template across tax courts. To identify the main structural headings of each judgment (in particular the section ``Motivi della decisione'', which contains the judge's reasoning) we relied on a regex-based matcher. Headings are not entirely uniform across decisions, so the matcher succeeds only on a subset of documents.

Before extraction we applied two filters intended to remove documents that do not contain legal issues and would therefore introduce noise into the subsequent analysis. First, we discarded judgments whose ``Motivi della decisione'' section is shorter than $680$ characters when the section was identified by the aforementioned matcher, and judgments whose total length is below $2{,}300$ characters when it was not: in both cases the text is too short to contain any interesting reasoning. Second, we filtered out judgments containing variants of ``estinzione del giudizio'' or ``cessata materia del contendere'', as these expressions signal that the proceedings were terminated or that the case had become moot, so that the decision does not resolve any issue on the merits. Of the $401{,}349$ judgments retrieved, $328{,}605$ ($81.9\%$) passed both filters and were processed by the extraction pipeline described in Section~\ref{sec:LLM_extraction}.
For the manual validation reported in Section~\ref{sec:validation_method}, we drew a subsample of $50$~judgments uniformly at random from the $328{,}605$ judgments in the final structured corpus.

\subsection{Overview of the extraction pipeline}
\begin{figure}
    \centering
    \includegraphics[width=\linewidth]{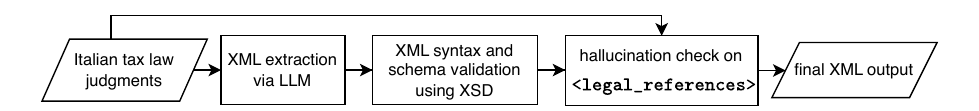}
    \caption{Stylized visualization of the extraction and verification pipeline.}
    \label{fig:flowchart_extraction}
\end{figure}

The overall extraction process can be represented with the flow chart in Figure \ref{fig:flowchart_extraction}. 
The text of each judgment is passed to LLM together with a prompt; an XML file, containing the structured information, is produced in output.
To verify that the XML outputted by the LLM conforms to what one asked for, the output is validated using an XSD (schema) file that specifies the fields that must be present and the values they are allowed to take.
The aim of the rest of the pipeline is to eliminate hallucinations from elements of \verb|<legal_references>|. To do so, the pipeline extracts and converts to a standardized format all references present in the text of the judgment. This is accomplished with the aid of the Linkoln \citep{linkoln} library. The same standardization is applied to references in the XML. Then the two sets are compared and references found in the XML but not in the judgment are marked as potential hallucinations and eliminated.

\subsection{LLM extraction}
\label{sec:LLM_extraction}
In this section we detail the XML format extraction procedure. 
To carry out the extraction we used the DeepSeek V3-0324 model \citep{liu2024deepseek}. After informally reviewing several LLMs for the issue extraction task, we chose DeepSeek because among the models achieving a satisfactory performance, it was the one with the best balance between capabilities and price (\$0.20/M input tokens, \$0.77/M output tokens). In contrast, top-range models can cost up to twenty times as much: for example Claude Sonnet 4.5 was priced at \$3/M input tokens and \$15/M output tokens. Many of the potential uses outlined in Section~\ref{sec:design_pressures} can only be achieved if a large number of judgments is processed. In our case, this amounts to processing several hundred thousand documents, and at this scale a ten-fold increase in price is very relevant to the feasibility of the extraction. The title-correction step described below uses the slightly newer DeepSeek V3.1 model (\$0.21/M input tokens, \$0.79/M output tokens). At list prices, the per-document cost amounts to approximately \$0.0032 for the main extraction with V3-0324 and \$0.00034 for the title-correction step with V3.1; processing the full corpus of $328{,}605$ judgments cost approximately \$1{,}150 in total.

The pre-processed corpus described in Section~\ref{sec:dataset} is the input to the extraction pipeline. We detail below the prompting strategy and a post-extraction correction step.
\begin{enumerate}
    \item If the ``Motivi della decisione'' section identified during preprocessing (Section~\ref{sec:dataset}) is shorter than $4100$ characters, the XML was extracted using prompt \ref{lst:prompt}, followed by the entire text of the judgment.\footnote{When the section ``Motivi della decisione'' is not present we default to the single step prompt.} Otherwise, a two-step approach was used. In the first step, prompt \ref{lst:prompt_multistep_questioni} is passed together with the judgment's text to the LLM. This first prompt identifies the issues within the judgment and extracts the \verb|<text>,<summary>, <issue_outcome>| for each issue. 
    In the second step, the following procedure is repeated for all issues. We pass prompt \ref{lst:prompt_multistep_reasoning}, the fields \verb|<text>| and \verb|<issue_outcome>| of the issue, and the text of the judgment to the LLM. As output we collect the \verb|<factual_premises>, <legal_references>, <citation_reason>, <judge_reasoning>| fields. The new fields are then merged to ones extracted in step 1, obtaining the complete XML format, as presented in \ref{lst:prompt}.
    We chose to split the extraction of the issues because the LLM was not producing satisfactory outputs on longer judgments containing several issues. We remarked that the LLM had a 'preferred' output length range and that in the case of judgments requiring longer XMLs, the single prompt approach led to too concise outputs and missing issues. We believe that using more powerful LLMs the single prompt approach would suffice.
    In our analysis, we relied exclusively on zero-shot prompting. To implement few-shot prompting one would need to provide the text of several judgments along with the extracted issues. Given that judgments are usually several pages long, this is unfeasible. Additionally, there is the risk of biasing the extraction towards the topics selected in the example judgments.
    \item Early experiments with this setting revealed that sometimes the \verb|<text>| field was not in the "Whether + clause" form, with the LLM mistakenly using a direct interrogative form, or other incorrect forms. For example instead of writing \verb|<text>Whether the taxpayer correctly calculated the VAT due </text>| the LLM could sometime formulate the issue as \verb|<text>Did the taxpayer correctly calculate the VAT due?</text>|.
    This problem affected about 10\% of the extracted issues.
    We therefore attempted to correct these fields automatically using few-shot prompting. Prompt \ref{lst:issue_text_correction}, followed by each of the \verb|<text>| fields, was fed into the LLM (DeepSeek V3.1 for this step). 
    Once again we expect this step not to be necessary if one is using a more powerful LLM. After this additional processing basically all issues were in the correct form.
\end{enumerate}
The steps we described illustrate in detail the procedure employed to extract the XML. We stress that irrespectively of whether prompt \ref{lst:prompt} or \ref{lst:prompt_multistep_questioni} and \ref{lst:prompt_multistep_reasoning} were used, the XML format is always the same. The LLM was run at zero temperature and with a fixed random seed using API calls to third party providers. Despite these measures the output of the model was not deterministic, yielding different results on identical inputs. We quantify the impact of this non-determinism in Appendix~\ref{app:sensitivity}, where we run the pipeline twice on the validation set and compare the two runs with the same set-based metrics used for the human validation.
\subsection{Verification procedure}
We now move to describing the post extraction checks we performed on the output of the LLM. 
The checks we are able to carry out automatically involve three aspects
\begin{enumerate}
    \item Syntax of the XML. Occasionally (in less than 3\% of cases) the LLM misses an XML closing tag '\verb|>|' or makes another syntax error. This causes the XML parser to throw an error. In this case the corresponding judgment is rerun through the LLM until the XML is well formed.
    \item Schema conformity of the XML. The prompt specifies a particular structure for the output XML, with constraints on elements and attributes (e.g. the \verb|type| attribute in \verb|legal_references| can only take values in \verb|leg,caselaw, princ, admin_pract|. To check that the LLM respected these constraints we employ an XML Schema Definition (XSD) file. This file allows to express the  constraints in a machine-readable form. One can then automatically verify the conformity of the XMLs. The fraction of non conformal XMLs is below 5\%. We rerun the corresponding judgments through the LLM pipeline until the output XML passes the XSD check.
    \item The most onerous check we carry out is on the presence of hallucinations within \verb|<legal_references>|. LLMs can in fact make up issues, facts, and legal references that are not present in the judgment. Ideally one would like to put in place some system that allows to detect and then eliminate hallucinations automatically. For free-text fields, this is quite hard to accomplish: essentially one would need an additional LLM to evaluate the extraction outputs (with the risk that also this second model hallucinates). A more rigorous approach can be taken concerning legal references, since their presence in the text of the judgment can be more readily verified. Citations can be of four types: legislation, case law, administrative practice, and principle of law. Let us focus for the moment on non-principle citations. We use several methods in succession to determine whether a citation is present in the text. 
    \begin{enumerate}
        \item Textual match. If the reference text (contained in the \verb|ref| attribute) appears in the judgment's text, then the citation is verified. To make matches more likely, we convert text to lowercase and remove spaces and non alphanumeric characters.
        \item Unique Resource Identifier (URI) based matching. If possible, we assign a URI (for example ECLI or CELEX) to the legal reference using Linkoln. Then, we assign URIs to all legal citations appearing in the judgment. Afterwards, we check whether the URI of the reference in the XML is among those found in the judgment. 
        \item Number-year matching. Legal documents can usually be uniquely identified by the following: type of document (e.g. judgment, decree, law, order), the issuing authority (e.g. regional tax court, president of the republic), number, and year. It becomes significantly easier to match references based exclusively on number and year as these are numerical values and therefore not subject to abbreviation. Given the number and year of a reference in the XML, we first look for the number in the judgment. If this is found we look for the year in a window of $\pm 100$ characters from the number. If also the year is found, the reference is considered verified.\footnote{This approach potentially misses hallucinated references that have the same number and year as the ones in the judgment but different authority. This is however very rare.} 
    \end{enumerate}
    The above pipeline is unsatisfactory for citations of type \verb|princ|,  since principles of law (e.g. principle of proportionality, principle of legality) are usually not associated with a URI or a combination of number + year. On top of this, we notice that LLMs do not have a clear understanding of what constitutes a principle, and tend to put other legal concepts (e.g. taxability of building areas, fiscal residence) under this category.
    We therefore employ a different approach. Based on the whole corpus of extracted issues (i.e. the $328{,}605$ processed files), minus the 50 test judgments, the experts jointly produced a list of 182 allowed legal principles. For each principle in the list a canonical form was established together with a series of possible variations (e.g. principle of legal certainty, principle of certainty of the law). Each reference of type \verb|princ| is kept if both of the following conditions are met.
    (1) there is a textual match in the judgment (computed in the same way as with other citations) and (2) the reference matches one of the principles in the list.
    
    The flowchart in figure \ref{fig:hallu_check_flowchart} illustrates the logic of the whole hallucination checker. This procedure is applied to each \verb|<item>| in \verb|<legal_references>|\footnote{It is possible, that one item contains more than one reference. In this case we remove the item only if all references identified within the item are recognized as hallucinated.}
\begin{figure}
    \centering
    \includegraphics[width=1.0\linewidth]{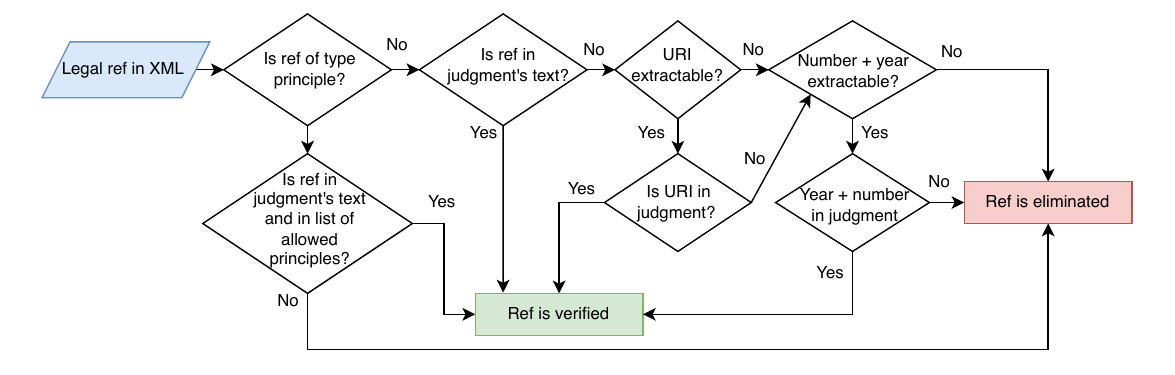}
    \caption{Flowchart of the hallucination check for items in <legal\_references>. }
    \label{fig:hallu_check_flowchart}
\end{figure}
\end{enumerate}

\section{Validation Protocol}
\label{sec:validation_method}

To assess the quality of the extraction pipeline, we conducted a manual validation of the LLM outputs on the $50$-judgment subsample described in Section~\ref{sec:dataset}. Two tax-law experts (PhD in tax law), hereafter $A_1$ and $A_2$, served as annotators.

\subsection{Development of the annotation guidelines}
Prior to the validation task, $A_1$ and $A_2$ jointly developed a set of annotation guidelines through an iterative process consisting of two preparatory rounds, both carried out on judgments disjoint from the test set. In a first round, the annotators jointly read and discussed $10$~judgments while examining the corresponding LLM outputs, in order to identify recurring ambiguities and edge cases (e.g.\ partially overlapping issues, citations to general legal principles, judgments concerning exclusively costs) and to converge on a common scoring scheme. In a second round, they independently annotated $10$~additional judgments under blind conditions, then compared their outputs, resolved disagreements through discussion, and used the residual divergences to produce the final version of the guidelines. The resulting document fixes both the scope of the annotation task and the criteria for each metric; the full guidelines are reported in Appendix~\ref{app:scoring_criteria}.

The central outcome of this phase was the adoption of a working definition of a \emph{legal issue} (\emph{questione giuridica}): a legal question explicitly decided by the judge in the reasoning section of the judgment, requiring the interpretation or application of a legal source, and supported by argumentative content sufficient to fill the structured fields of the XML format. This definition focuses on the reasoning of the judge as it emerges from the argumentative part of the decision, rather than on the requests of the parties, and was informed by the literature in civil procedural law and the philosophy of law \citep{villa2023graduazione,Ricci1987Accertamento, liebman2007manuale}. The choice is consistent with the intended use of the extracted format, which is to support access to legal issues as they are addressed and resolved by judges. The definition deliberately admits a small residual margin of interpretation: as the inter-annotator agreement results in Section~\ref{sec:results} show, even two experts working under the same guidelines may segment a judgment into slightly different but largely overlapping sets of issues.

\subsection{Annotation procedure}
For each judgment, the annotator carried out the following four steps, in order. All annotations were recorded in a spreadsheet with one row per extracted issue.

\begin{enumerate}
    \item \textbf{Independent issue identification.} Without consulting the LLM output, the annotator read the full text of the judgment and compiled a list of the legal issues present, formulating each as a \emph{Whether} + clause (Italian: \emph{Se + congiuntivo}), that is,  in the same form used by the LLM in the \verb|<text>| field.

    \item \textbf{Issue alignment and scoring.} The annotator then examined the LLM-extracted issues and aligned them with their own list, on the basis of content rather than literal wording. This step partitions the LLM issues into \emph{present} (those matching an issue in the annotator's list) and \emph{absent} (those that do not correspond to any issue the annotator considered present in the judgment). The alignment also identifies issues in the annotator's list that the LLM failed to extract. Across the entire test set, fewer than five alignments were considered ambiguous and were resolved by discussion between the two annotators. For each LLM issue marked as present, the annotator additionally rated, on a $1$--$5$ Likert scale, the appropriateness of its \emph{level of generality} (neither too narrow nor too broad for a caselaw research) and the quality of its \emph{legal language}.

    \item \textbf{Citation evaluation.} For each present issue, and adopting the LLM's formulation of that issue, the annotator independently compiled the list of legal references they considered relevant to its resolution, without looking at \verb|<legal_references>|. This list was then aligned with the LLM-extracted citations. 
    The annotator further scored the \verb|<citation_reason>| entries on a binary scale ($1$ if the stated reason corresponds to the use the judge makes of the citation in the decision, $0$ otherwise).

    \item \textbf{Free-text field scoring.} Finally, the annotator scored the remaining textual fields. The \verb|<summary>| field was rated along four dimensions on $1$--$5$ Likert scales: \emph{correctness} (absence of elements not present in the source), \emph{form} (coherence, readability, legal terminology), \emph{completeness} (coverage of the key points of the issue), and overall \emph{satisfaction}. The \verb|<factual_premises>|, \verb|<judge_reasoning>|, and \verb|<issue_outcome>| fields each received a single overall satisfaction score on the same scale.
\end{enumerate}

\subsection{Two-phase rollout}
The $50$ judgments were annotated in two phases. In the \emph{first phase}, both $A_1$ and $A_2$ independently annotated the same $20$~judgments under double-blind conditions. This shared subset serves the  purpose of supporting the computation of inter-annotator agreement (IAA). Where IAA is high, deviations of the LLM output from the human annotations can plausibly be attributed to model error; where IAA is lower, part of the gap should instead be read as an irreducible disagreement between experts on a task that even humans would struggle to resolve consistently. In the \emph{second phase}, the remaining $30$~judgments were split evenly between the two annotators, each annotating $15$~judgments independently. Each annotator therefore produced annotations for a total of $35$~judgments, with $20$ in common.

\subsection{Evaluation metrics}
For both issue extraction and citation extraction, the LLM output and the annotator output are sets of items (issues or citations), and we evaluate them with standard precision (P), recall (R), and F1. To define these unambiguously, fix an annotator $A \in \{A_1, A_2\}$. Then
\[
\mathrm{P}_{\mathrm{LLM}|A} = \frac{|\mathrm{LLM} \cap A|}{|\mathrm{LLM}|}, \qquad
\mathrm{R}_{\mathrm{LLM}|A} = \frac{|\mathrm{LLM} \cap A|}{|A|},
\]
i.e.\ the annotator's set is treated as ground truth. The same formulas, with $A_1$ and $A_2$ in turn playing the role of ground truth, are used to compute IAA on the shared subset; the two directions ($A_1$ as ground truth for $A_2$ and vice versa) yield different precision and recall but the same F1.

All set-level scores in the main text are computed by pooling items across the relevant set of judgments before computing the metric. As a robustness check, Appendix~\ref{app:macro} reports these metrics averaged per judgment rather than pooled, and Appendix~\ref{app:union_inter} against a combined-annotator ground truth. To quantify the sampling uncertainty of these pooled scores, we accompany the main tables with $95\%$ confidence intervals computed with a non-parametric cluster bootstrap at the judgment level: the judgments, with all their issues and citations, are resampled with replacement ($B{=}10{,}000$), the pooled metric is recomputed on each replicate, and we report the $2.5$th--$97.5$th percentile interval. Resampling entire judgments preserves the dependence among items belonging to the same decision; the judgment is also the unit at which the validation sample was drawn. Intervals are shown in brackets in the corresponding tables. For the qualitative Likert scores, we report exact agreement, within-one-point agreement, and Gwet's $\mathrm{AC2}$ coefficient with quadratic weights \citep{gwet2014handbook}, the latter being designed to remain stable in the high-prevalence regime in which annotators agree on most items. We do not adopt Cohen's $\kappa$ as the primary agreement statistic because $\kappa$ is known to collapse under marginal skew \citep{feinstein1990high,wongpakaran2013comparison}, a condition that holds for several of our dimensions; for completeness, $\kappa$ values are reported in Appendix~\ref{app:kappa} together with a brief discussion of their interpretation here.

For the hallucination filter, we additionally report its specificity (the fraction of valid citations correctly retained), its precision and recall as a binary classifier, and the residual hallucination rate among the citations it keeps. These are defined and discussed in Section~\ref{sec:results_hallucination}.

After validation, hallucinated issues (i.e.\ LLM-extracted issues that the annotator marked as absent) are dropped from the analysis of all other fields, since their associated facts, references, reasoning, and summary have no annotator counterpart.

\section{Results}
\label{sec:results}

We organize the results as follows. Section~\ref{sec:results_issues} reports IAA and LLM-vs-annotator agreement on the extraction of legal issues; Section~\ref{sec:results_citations} does the same for legal citations (\verb|<legal_references>| and \verb|<citation_reason>|); Section~\ref{sec:results_hallucination} evaluates the hallucination filter as a stand-alone component; and Section~\ref{sec:results_scores} reports the qualitative Likert scores on the free-text fields.
\subsection{Legal issues extraction}
\label{sec:results_issues}
We start by considering the average number of legal issues per judgment extracted by the LLM and by each annotator. From the results, displayed in Table~\ref{tab:issue_counts}, a consistent ordering emerges: $A_1$ identifies more issues than $A_2$, who in turn identifies more than the LLM, both on the shared subset and on the full sets. The LLM extracts on average $1.5$~issues per judgment, against roughly $2$ for the human annotators, indicating a tendency to under-extract.

\begin{table}[!htbp]
\centering
\caption{Average number of issues identified per judgment (total issues in parentheses).}
\label{tab:issue_counts}
\begin{tabular}{l cc}
\toprule
 & Shared ($N{=}20$) & Full set \\
\midrule
$A_1$  & 2.35 (47) & 2.09 (73) \quad $N{=}35$ \\
$A_2$  & 1.85 (37) & 1.77 (62) \quad $N{=}35$ \\
LLM & 1.50 (30) & 1.56 (78) \quad $N{=}50$ \\
\bottomrule
\end{tabular}
\end{table}
Next, we consider the inter annotator agreement on the shared set ($N=20$).  The two annotators show very strong agreement on issue identification: F1${=}88.1\%$ on the shared set (Table~\ref{tab:issue_pairwise}, top rows). The residual disagreement, despite shared guidelines, sets the natural ceiling against which the LLM should be evaluated.
Table~\ref{tab:issue_pairwise} also shows the performance of the LLM's extraction against the ground truth set by the experts. The LLM achieves high precision against both annotators ($93.3\%$), meaning that the model rarely produces an issue that the annotator considers absent (only two such cases on $N{=}20$, confirmed independently by both annotators). Recall is lower and varies with the reference annotator: the LLM recovers about $60\%$ of $A_1$'s issues but $76\%$ of $A_2$'s. The pattern is confirmed on the full $N{=}35$ sets (Table~\ref{tab:issue_n35}), where precision remains above $94\%$ while recall ranges from $69.9\%$ (against $A_1$) to $82.3\%$ (against $A_2$). Across all comparisons, the dominant source of error is under-extraction rather than hallucination. This profile is partly by design. By asking the model for the ``one or two'' main issues in the prompt, we deliberately trade recall for precision. Hallucinated issues are indeed rare in absolute terms: on the whole dataset ($N=50$), only $4$ of the $78$ LLM-extracted issues were marked as absent by either annotator. 

These $4$ issues are excluded from the subsequent analyses. 

\begin{table}[!htbp]
\centering
\caption{Pairwise agreement on issue extraction ($N{=}20$ shared judgments). Each row treats the second item as ground truth. In brackets: $95\%$ judgment-level bootstrap confidence intervals.}
\label{tab:issue_pairwise}
\begin{tabular}{l ccc}
\toprule
Comparison \\(ref.\ system $\mid$ ground truth) & P & R & F1 \\
\midrule
$A_1$ $\mid$ $A_2$          & 78.7 [65.6, 95.7] & 100.0 & 88.1 [79.2, 97.8] \\
$A_2$ $\mid$ $A_1$          & 100.0 & 78.7 [65.6, 95.7] & 88.1 [79.2, 97.8] \\
LLM $\mid$ $A_1$         & 93.3 [83.9, 100.0] & 59.6 [42.0, 84.8] & 72.7 [56.4, 91.2] \\
LLM $\mid$ $A_2$         & 93.3 [83.9, 100.0] & 75.7 [55.8, 100.0] & 83.6 [67.5, 100.0] \\
\bottomrule
\end{tabular}
\end{table}

\begin{table}[!htbp]
\centering
\caption{LLM issue extraction evaluated against each annotator's full set ($N{=}35$). In brackets: $95\%$ judgment-level bootstrap confidence intervals.}
\label{tab:issue_n35}
\small
\begin{tabular}{l cc}
\toprule
 & LLM $\mid$ $A_1$ & LLM $\mid$ $A_2$ \\
\midrule
Precision  & 94.4 [87.9, 100.0] & 94.4 [88.0, 100.0] \\
Recall     & 69.9 [54.2, 87.5] & 82.3 [67.9, 96.2] \\
F1         & 80.3 [67.6, 92.2] & 87.9 [77.1, 97.1] \\
\bottomrule
\end{tabular}
\end{table}

\subsection{Legal citation extraction}
\label{sec:results_citations}

All citation results in this section refer to the LLM output \emph{after} the hallucination filter has been applied. Unless otherwise noted, citation metrics are restricted to the issues marked as present by the relevant annotator.

Table~\ref{tab:cit_counts} reports the average number of legal references per present issue. We remark that the LLM extracts about the same number of references per issue as the annotators.
\begin{table}[!htbp]
\centering
\caption{Average number of legal references identified per present issue (total citations in parentheses).}
\label{tab:cit_counts}
\begin{tabular}{l cc}
\toprule
 & Shared ($N{=}20$) & Full set \\
\midrule
$A_1$  & 2.39 (67)  & 2.90 (148) \quad $N{=}35$ \\
$A_2$  & 2.46 (69)  & 2.82 (144) \quad $N{=}35$ \\
LLM & 2.57 (72)  & 3.08 (228) \quad $N{=}50$ \\
\bottomrule
\end{tabular}
\end{table}

The two annotators show near-perfect agreement on the identification of citations relevant to each issue (F1${=}97.1\%$; Table~\ref{tab:cit_pairwise}). This is coherent with the less discretionary nature of the task.
Against this reference, the LLM achieves F1 around $74\%$, with precision around $72\%$ and recall around $75$--$77\%$. Notice that most of the LLM citations identified as irrelevant by the annotators are still present in the judgment text (and so are not hallucinations); they are simply not deemed relevant to the resolution of the specific issue. We remark that running Linkoln on the ``Motivi'' section of the judgment is a competitive baseline on recall (Appendix~\ref{app:linkoln_baseline}); the LLM's distinctive contribution is attributing each citation to the specific issue it resolves.
\begin{table}[!htbp]
\centering
\caption{Pairwise agreement on citation extraction ($N{=}20$ shared judgments, $28$~present issues). Each row treats the second item as ground truth. In brackets: $95\%$ judgment-level bootstrap confidence intervals.}
\label{tab:cit_pairwise}
\small
\begin{tabular}{l ccc}
\toprule
Comparison (ref.\ system $\mid$ ground truth) & P & R & F1 \\
\midrule
$A_1$ $\mid$ $A_2$        & 98.5 [95.7, 100.0] & 95.7 [89.9, 100.0] & 97.1 [94.1, 100.0] \\
$A_2$ $\mid$ $A_1$        & 95.7 [89.9, 100.0] & 98.5 [95.7, 100.0] & 97.1 [94.1, 100.0] \\
LLM $\mid$ $A_1$       & 72.2 [53.8, 90.4] & 77.6 [68.7, 88.9] & 74.8 [62.8, 85.3] \\
LLM $\mid$ $A_2$       & 72.2 [53.8, 90.4] & 75.4 [68.3, 85.2] & 73.8 [62.4, 83.3] \\
\bottomrule
\end{tabular}
\end{table}
On each annotator's full $N{=}35$ set (Table~\ref{tab:cit_n35}), precision is stable at $73\%$ and recall is in the range  $70$--$80\%$, depending on the annotator.

\begin{table}[!htbp]
\centering
\caption{LLM citation extraction evaluated against each annotator's full set ($N{=}35$). In brackets: $95\%$ judgment-level bootstrap confidence intervals.}
\label{tab:cit_n35}
\small
\begin{tabular}{l cc}
\toprule
 & LLM $\mid$ $A_1$ & LLM $\mid$ $A_2$ \\
\midrule
Precision & 72.7 [60.5, 85.8] & 73.9 [60.1, 86.8] \\
Recall    & 70.3 [60.9, 84.2] & 80.6 [73.4, 88.1] \\
F1        & 71.5 [62.8, 80.7] & 77.1 [66.9, 85.3] \\
\bottomrule
\end{tabular}
\end{table}

\paragraph{Faithfulness of citation reasons.}
For each LLM-extracted citation, the annotators also scored the corresponding \verb|<citation_reason>| field on a binary scale (Section~\ref{sec:validation_method}, step~3): $1$ if the stated reason corresponds to the use that the judge makes of the citation in the decision, $0$ otherwise. The natural summary statistic is the fraction of LLM-extracted reasons judged faithful, pooled across issues. Table~\ref{tab:cit_reason} reports this fraction on the shared subset ($N=20$, $49$~scored reasons). The two annotators agree on the count of faithful reasons in $96.4\%$ of issues. On the full corpus ($N=50$), the LLM achieves a faithfulness rate of $80\%$, indicating that four out of five citation reasons produced by the model correctly summarize the role that the corresponding citation plays in the decision.

\begin{table}[!htbp]
\centering
\caption{Citation-reason faithfulness: fraction of LLM-extracted citation reasons judged faithful by the annotators (means). Shared subset: $N{=}20$, $28$~present issues, $49$~reasons. Full corpus: $N=50$.}
\label{tab:cit_reason}
\begin{tabular}{l c}
\toprule
 & Value \\
\midrule
\multicolumn{2}{l}{\emph{Mean faithfulness rate (pooled)}} \\
\quad $A_1$, shared    & 81.6 \\
\quad $A_2$, shared    & 75.5 \\
\quad LLM, full corpus &  80.8\\
\bottomrule
\end{tabular}
\end{table}

\subsection{Hallucination filter}
\label{sec:results_hallucination}

We now evaluate the hallucination detection step as a stand-alone component, on the full corpus ($N{=}50$, $264$~initially extracted citations). Before and after applying the filter, the annotators recorded whether each reference actually appears in the judgment text; this provides the ground truth against which the filter is assessed. In the case of principles, we consider hallucinated references that do appear in the text but that clearly do not correspond to a recognized legal principle.

Out of 264 total citations, 31 were found to be hallucinated. Of the 31, 20 were of type principle showing that the LLM performs particularly poorly on this category, which over the test set of 50 judgments only counts 29 references, giving a hallucination rate of 69\% on principles.\footnote{One can break this further down into concepts that appear in the judgment's text but are not principles (12/20) and proper principles that are not cited in the judgment (8/20).}. In comparison, the hallucination rate for other references is $11/235=4.7\%$.
The filter removes $36$ of the $264$ extracted citations (29 truly hallucinated and 7 not), reducing the output set to $228$. Table~\ref{tab:halluc_rates} shows the net effect: the overall hallucination rate drops from $11.7\%$ ($95\%$ CI $[6.6, 17.8]$) to $0.9\%$ ($[0.0, 2.3]$). Crucially, this comes at a very low cost in valid citations: of the $233$ non-hallucinated citations, only $7$ ($3\%$) are incorrectly removed.
\begin{table}[!htbp]
\centering
\caption{Effect of the hallucination filter ($N{=}50$). In brackets: $95\%$ judgment-level bootstrap confidence intervals.}
\label{tab:halluc_rates}
\small
\begin{tabular}{l cccc}
\toprule
 & Citations & Hallucinated & Halluc.\ rate & Valid lost \\
\midrule
Before filter & 264 & 31 & 11.7\% [6.6, 17.8] & --- \\
After filter  & 228 & 2 & 0.9\% [0.0, 2.3] & 7 (3.0\% [1.1, 5.3] of valid) \\
\bottomrule
\end{tabular}
\end{table}
The filter's full confusion matrix is reported in Table~\ref{tab:halluc_confusion}.
\begin{table}[!htbp]
\centering
\caption{Confusion matrix of the hallucination detector ($N{=}50$, $264$~citations). Specificity${=}97.0\%$.}
\label{tab:halluc_confusion}
\begin{tabular}{l cc c}
\toprule
 & Flagged & Not flagged & Total \\
\midrule
Actual hallucination     &  29  & 2  & 31 \\
Not hallucination        &  7  & 226 & 233 \\
\midrule
Total                    & 36  & 228 & 264 \\
\bottomrule
\end{tabular}
\end{table}
There are only two residual hallucinations, both of type \verb|caselaw|. These correspond to made up references whose year and number match some combination of year and number found in the judgment, but the issuing authority is incorrect.
\subsection{Qualitative scores on free-text fields}
\label{sec:results_scores}

Table~\ref{tab:quality_scores} reports inter-annotator agreement on the $1$--$5$ Likert scales, computed on the shared subset ($N{=}20$, $28$~present issues), alongside the average LLM score on each dimension, computed on the full corpus ($N{=}50$, with shared judgments contributing the mean of the two annotators' scores).

Agreement is uniformly high across all dimensions. The exact agreement ranges from $64.3\%$ (generality of the issue) to $100\%$ (outcome satisfaction); the agreement within a point is at least $89\%$ in every dimension and reaches $100\%$ in legal language, form of the summary and outcome satisfaction. The chance-corrected Gwet's $\mathrm{AC2}$ never drops below $0.93$. The two annotators thus converge on a shared expert evaluation of the LLM's free-text output. Cohen's $\kappa$ values for the same data are reported and discussed in Appendix~\ref{app:kappa}. We remark, however, that in our saturated regime, where most items receive the same rating $\kappa$ is unstable and not informative on its own. 

\begin{table}[!htbp]
\centering
\caption{Inter-annotator agreement on Likert dimensions ($N{=}20$, $28$~present issues) and average LLM scores ($N{=}50$). EA: exact agreement; W1: within-one-point agreement; AC2: Gwet's $\mathrm{AC2}$ with quadratic weights.}
\label{tab:quality_scores}
\small
\begin{tabular}{l cc ccc c}
\toprule
 & \multicolumn{5}{c}{Inter-annotator ($N{=}20$)} & LLM avg \\
\cmidrule(lr){2-6}
Dimension & Mean $A_1$ & Mean $A_2$ & EA & W1 & AC2 & ($N{=}50$) \\
\midrule
Issue (\verb|<text>|) generality        & 4.75 & 4.32 & 64.3\%  & 92.9\%  & 0.93 & 4.61 \\
Issue (\verb|<text>|) legal language          & 4.89 & 4.79 & 89.3\%  & 100\%   & 0.99 & 4.84 \\
\verb|<summary>| correctness     & 4.79 & 4.86 & 85.7\%  & 96.4\%  & 0.97 & 4.74 \\
\verb|<summary>| form            & 4.86 & 4.93 & 78.6\%  & 100\%   & 0.98 & 4.82 \\
\verb|<summary>| completeness    & 4.86 & 4.86 & 85.7\%  & 92.9\%  & 0.97 & 4.85 \\
\verb|<summary>| satisfaction    & 4.89 & 4.82 & 85.7\%  & 92.9\%  & 0.97 & 4.77 \\
\verb|<factual_premises>| satisfaction       & 4.79 & 4.54 & 82.1\% & 89.3\% & 0.94 & 4.68 \\
\verb|<judge_reasoning>| satisfaction & 4.93 & 4.82 & 89.3\%  & 92.9\%  & 0.98 & 4.78 \\
\verb|<issue_outcome>| satisfaction    & 5.00 & 5.00 & 100\%   & 100\%   & 1.00 & 4.89 \\
\bottomrule
\end{tabular}
\end{table}
\section{Discussion and Conclusion}
\label{sec:conclusion}

We have presented an extraction pipeline that decomposes court judgments into autonomous \emph{issues} represented in a structured XML schema grounded in IRAC and in the legal syllogism. The pipeline is paired with an automatic hallucination-detection step that compares LLM-extracted legal references against the judgment text after normalization to standard identifiers (urn:nir, CELEX, ECLI). The pipeline was developed and validated on a corpus of Italian first- and second-instance tax-court judgments, a sector characterized by hypertrophic case-law production, heterogeneous drafting styles (partly due to the involvement of non-career honorary judges), and a thematic complexity that draws on civil, criminal, and commercial law.

The validation on $50$ judgments, doubly annotated by two tax-law experts on a shared subset of $20$, supports four main claims. First, issue extraction is high-precision and moderate-recall: of the $78$ issues extracted by the LLM, only $4$ were marked as absent by either annotator, while the model recovers between $69.9\%$ and $82.3\%$ ($N=35$) of the issues identified by the human annotators. Under-extraction is the dominant error. Second, citation extraction shows moderate precision and recall, both above $70\%$. Third, the hallucination filter effectively removes most fabricated references: at the cost of incorrectly removing only $7$ of $233$ valid citations ($3.0\%$), it eliminates $29$ of $31$ hallucinated references, bringing the residual hallucination rate from $11.7\%$ to $0.9\%$.  The type of reference most subject to hallucinations are principles, with a hallucination rate of $69\%$. We attribute this to LLMs having little knowledge of what constitutes a legal principle and therefore tending to include irrelevant elements.
Fourth, the qualitative Likert scores on the free-text fields are uniformly above $4.6/5$ on the summary, judge's reasoning, and issue outcome, with slightly lower (yet strong) scores on factual premises and on the level of generality of the issue formulation, due to both being intrinsically more discretionary tasks.

A methodological observation cuts across these results. The two annotators converged on a shared expert evaluation across nearly all tasks, with the lowest (yet still strong) agreement on issue identification itself. The choice of how to segment a decision into autonomous questions retains an irreducible subjective margin: part of the gap between LLM and humans on issue extraction is therefore a property of the task.

Beyond the specific results, the contribution we wish to emphasize is methodological. An issue-based, IRAC-aligned representation, validated by experts and equipped with citation-level hallucination control, offers a template for structured extraction over case law that is independent of the specific jurisdiction and of the specific LLM. Our validation effort demonstrates that current LLMs are able to reliably instantiate the proposed XML format from judgments.
This opens up a range of applications, from issue-level retrieval, to the construction of citation networks whose edges reflect actual reasoning rather than incidental mentions, to the creation of large-scale datasets that provide a structured representation of the reasoning contained in case law.

\paragraph{Limitations and future work}
The validation rests on $50$ judgments and two annotators, which, while consistent with the cost of expert annotation, limits the tightness of the reported estimates. The results are tied to one model (DeepSeek V3) and one corpus (Italian tax-court judgments); LLM outputs were moreover non-deterministic even at zero temperature and fixed random seed, in line with recent findings~\citep{blair2025llms}, so replication on other models and other branches of law is needed to confirm generality.

Several directions naturally extend the present contribution. First, one can apply the same format to other types of judgments, both within Italian jurisprudence and across jurisdictions, to test the generality of an issue-based representation. A second direction is to exploit the \verb|<citation_reason>| field to enrich the description of cited documents, for instance by associating each piece of legislation with the list of issues that cite it and with the reasons for citation. A third, related direction is to leverage the structured textual fields and the citation network for retrieval, possibly combining issue-level vector representations with graph-based ranking algorithms such as PageRank computed on the resulting issue-to-issue citation graph.
\section{Acknowledgments}
P.V. and G.P. acknowledge support from UKRI FLF Scheme (No. MR/X023028/1)
\section{Code and Data Availability}
The validation code and data are available at \url{https://github.com/giovannipiccioli/legal_issues_extraction}. The repository contains the expert annotations, the raw and post-filter XML extractions of the 50 test judgments for both runs, the list of 182 canonical legal principles, and five notebooks that reproduce every table in Section~\ref{sec:results} and the appendices.

\bibliographystyle{plainnat}

\bibliography{references}  

@article{linkoln,
  title={Improving public access to legislation through legal citations detection: the linkoln project at the Italian senate},
  author={Bacci, L. and Agnoloni, T. and Marchetti, C. and Battistoni, R.},
  journal={Knowledge of the Law in the Big Data Age},
  volume={317},
  pages={149--158},
  year={2019},
  publisher={IOS Press},
  doi={10.3233/FAIA190017}
}

@online{banca_dati_giurisprudenza_tributaria,
  title        = {Banca Dati della Giurisprudenza Tributaria},
  author       = {{Ministero dell'Economia e delle Finanze -- Dipartimento della Giustizia Tributaria}},
  year         = {2024},
  url          = {https://bancadatigiurisprudenza.giustiziatributaria.gov.it/},
  note         = {Accessed: 2026-04-29}
}

@software{pymupdf,
  title        = {PyMuPDF Documentation},
  author       = {{Artifex Software, Inc.} and McKie, Jorj X.},
  year         = {2026},
  url          = {https://pymupdf.readthedocs.io/en/latest/},
  note         = {Accessed: 2026-04-29}
}

@article{francesconi2010urn,
  title={URN-based Identification of Legal Acts: The Case of the Italian Senate},
  author={Francesconi, Enrico and Marchetti, Carlo and Pietramala, Remigio and Spinosa, Pierluigi},
  journal={Informatica e diritto},
  volume={19},
  number={1-2},
  pages={233--252},
  year={2010},
  note={No DOI available; article predates DOI assignment for this journal}
}

@article{adhikary2024case, title={A case study for automated attribute extraction from legal documents using large language models}, author={Adhikary, Subinay and Sen, Procheta and Roy, Dwaipayan and Ghosh, Kripabandhu}, journal={Artificial Intelligence and Law}, pages={1--22}, year={2024}, publisher={Springer}, doi={10.1007/s10506-024-09425-7}, note={Online-first pagination shown; later assigned to vol. 34, pp. 245--266} }

@article{kang2025automating, title={Automating IRAC analysis in Malaysian contract law using a semi-structured knowledge base}, author={Kang, Xiaoxi and Qu, Lizhen and Soon, Lay-Ki and Li, Zhuang and Trakic, Adnan}, journal={Artificial Intelligence and Law}, pages={1--44}, year={2025}, publisher={Springer}, doi={10.1007/s10506-025-09467-5} }

@article{pont2023legal, title={Legal summarisation through llms: The prodigit project}, author={Dal Pont, Thiago and Galli, Federico and Loreggia, Andrea and Pisano, Giuseppe and Rovatti, Riccardo and Sartor, Giovanni}, journal={arXiv preprint arXiv:2308.04416}, year={2023}, doi={10.48550/arXiv.2308.04416} }

@article{han2026legal, title={Legal citation prediction with LLMs: a comparative evaluation of instruction tuning, retrieval, and jurisdiction-specific pre-training on the AusLaw citation benchmark}, author={Han, Jiuzhou and Burgess, Paul and Shareghi, Ehsan}, journal={Artificial Intelligence and Law}, pages={1--35}, year={2026}, publisher={Springer}, doi={10.1007/s10506-026-09506-9} }

@article{yu2022legal, title={Legal prompting: Teaching a language model to think like a lawyer}, author={Yu, Fangyi and Quartey, Lee and Schilder, Frank}, journal={arXiv preprint arXiv:2212.01326}, year={2022}, doi={10.48550/arXiv.2212.01326} }

@article{benedetto2025legitbart, title={LegItBART: a summarization model for Italian legal documents}, author={Benedetto, Irene and La Quatra, Moreno and Cagliero, Luca}, journal={Artificial Intelligence and Law}, pages={1--31}, year={2025}, publisher={Springer}, doi={10.1007/s10506-025-09436-y} }

@article{habernal2024mining, title={Mining legal arguments in court decisions}, author={Habernal, Ivan and Faber, Daniel and Recchia, Nicola and Bretthauer, Sebastian and Gurevych, Iryna and Spiecker genannt D{\"o}hmann, Indra and Burchard, Christoph}, journal={Artificial Intelligence and Law}, volume={32}, number={3}, pages={1--38}, year={2024}, publisher={Springer}, doi={10.1007/s10506-023-09361-y} }

@incollection{grundler2025automated,
  title={Automated Extraction of Judicial Interpretative Formulas in EU Case Law on VAT},
  author={Grundler, Giulia and Santin, Piera and Fidelangeli, Alessia and Mignone, Rachele and Galli, Federico and Galassi, Andrea and Contissa, Giuseppe and Torroni, Paolo and others},
  booktitle={Frontiers in Artificial Intelligence and Applications},
  pages={294--299},
  year={2025},
  note={JURIX 2025; DOI could not be confirmed at time of verification}
}

@inproceedings{licari2023legal,
  title={Legal holding extraction from italian case documents using italian-legal-bert text summarization},
  author={Licari, Daniele and Bushipaka, Praveen and Marino, Gabriele and Comand{\'e}, Giovanni and Cucinotta, Tommaso},
  booktitle={Proceedings of the Nineteenth International Conference on Artificial Intelligence and Law},
  pages={148--156},
  year={2023},
  doi={10.1145/3594536.3595177}
}

@inproceedings{santin2023argumentation,
  title={Argumentation structure prediction in CJEU decisions on fiscal state aid},
  author={Santin, Piera and Grundler, Giulia and Galassi, Andrea and Galli, Federico and Lagioia, Francesca and Palmieri, Elena and Ruggeri, Federico and Sartor, Giovanni and Torroni, Paolo},
  booktitle={Proceedings of the Nineteenth International Conference on Artificial Intelligence and Law},
  pages={247--256},
  year={2023},
  doi={10.1145/3594536.3595174}
}

@inproceedings{westermann2025automated,
  title={Automated Mapping of Legal Criteria to the Texts of Adjudicatory Decisions Using LLMs},
  author={Westermann, Hannes and R. Walker, Vern and Savelka, Jaromir},
  booktitle={Proceedings of the Twentieth International Conference on Artificial Intelligence and Law},
  pages={219--228},
  year={2025},
  doi={10.1145/3769126.3769236}
}

@inproceedings{janatian2023text,
  title={From text to structure: Using large language models to support the development of legal expert systems},
  author={Janatian, Samyar and Westermann, Hannes and Tan, Jinzhe and Savelka, Jaromir and Benyekhlef, Karim},
  booktitle={Legal Knowledge and Information Systems: JURIX 2023: The Thirty-sixth Annual Conference, Maastricht, the Netherlands, 18--20 December 2023},
  pages={167--176},
  year={2023},
  organization={IOS Press},
  note={DOI could not be confirmed at time of verification}
}

@inproceedings{costa2024automated,
  title={Automated semantic annotation pipeline for brazilian judicial decisions},
  author={Costa, Melissa Zorzanelli and Robson, Dylan Faria and Vieira, Thiago Baiense Pe{\c{c}}anha and Bourguet, Jean-R{\'e}mi and Guizzardi, Giancarlo and Almeida, Jo{\~a}o Paulo A},
  booktitle={37th Annual Conference on Legal Knowledge and Information Systems, JURIX 2024},
  pages={226--238},
  year={2024},
  organization={IOS Press},
  doi={10.3233/FAIA241248}
}

@article{xu2023question,
  title={Question-Answering Approach to Evaluating Legal Summaries},
  author={Xu, Huihui and Ashley, Kevin},
  journal={arXiv preprint arXiv:2309.15016},
  year={2023},
  doi={10.48550/arXiv.2309.15016}
}

@inproceedings{garzo2025does,
  title={Does chatgpt understand the law? a case study on road homicide in italy},
  author={Garzo, Grazia and Palumbo, Alessandro},
  booktitle={JURIX 2025-38th International Conference on Legal Knowledge and Information Systems},
  pages={74--85},
  year={2025},
  doi={10.3233/FAIA251578}
}

@article{belfathi2023harnessing,
  title={Harnessing GPT-3.5-turbo for rhetorical role prediction in legal cases},
  author={Belfathi, Anas and Hernandez, Nicolas and Monceaux, Laura},
  journal={arXiv preprint arXiv:2310.17413},
  year={2023},
  doi={10.48550/arXiv.2310.17413},
  note={Also published as peer-reviewed paper at JURIX 2023 (DOI 10.3233/FAIA230964)}
}

@inproceedings{lombardi2023legal,
  title={Legal Text Segmentation Through Breakpoint Detection},
  author={Lombardi, Andrea and Alfano, Domenico and Abbruzzese, Roberto},
  booktitle={Legal Knowledge and Information Systems: JURIX 2023: The Thirty-sixth Annual Conference, Maastricht, the Netherlands, 18--20 December 2023},
  pages={227--236},
  year={2023},
  organization={IOS Press},
  note={DOI could not be confirmed at time of verification}
}

@article{mumford2025context,
  title={Context-Aware Citation Networks: A Human--AI Dataset, Analysis, and Tool},
  author={Mumford, Jack and Florimonte, Francesco and Atkinson, Katie and Dzehtsiarou, Kanstantsin},
  journal={Frontiers in Artificial Intelligence and Applications},
  year={2025},
  publisher={IOS Press},
  doi={10.3233/FAIA251584}
}

@incollection{zin2025court,
  title={From Court Decisions to Guiding Principles: Advancing Complex Legal Summarization with LLMs},
  author={Zin, May Myo and Satoh, Ken and Borges, Georg},
  booktitle={Legal Knowledge and Information Systems},
  pages={371--376},
  year={2025},
  publisher={IOS Press},
  note={Verified as a real JURIX 2025 paper; DOI could not be confirmed}
}

@inproceedings{arvin2025identifying,
  title={Identifying Legal Holdings with LLMs: A Systematic Study of Performance, Scale, and Memorization},
  author={Arvin, Chuck},
  booktitle={Proceedings of the Twentieth International Conference on Artificial Intelligence and Law},
  pages={404--408},
  year={2025},
  doi={10.1145/3769126.3769128}
}

@inproceedings{blair2025llms,
  title={Llms provide unstable answers to legal questions},
  author={Blair-Stanek, Andrew and Van Durme, Benjamin},
  booktitle={Proceedings of the Twentieth International Conference on Artificial Intelligence and Law},
  pages={425--429},
  year={2025},
  doi={10.1145/3769126.3769245}
}

@inproceedings{benedetto2023benchmarking,
  title={Benchmarking abstractive models for italian legal news summarization},
  author={Benedetto, Irene and Cagliero, Luca and Tarasconi, Francesco and Giacalone, Giuseppe and Bernini, Claudia},
  booktitle={Legal Knowledge and Information Systems: JURIX 2023: The Thirty-sixth Annual Conference, Maastricht, the Netherlands, 18--20 December 2023},
  pages={101--106},
  year={2023},
  organization={IOS Press},
  note={Pages corrected from 311--316 to 101--106; DOI could not be confirmed with certainty}
}

@inproceedings{jiang2023legal,
  title={Legal syllogism prompting: Teaching large language models for legal judgment prediction},
  author={Jiang, Cong and Yang, Xiaolei},
  booktitle={Proceedings of the nineteenth international conference on artificial intelligence and law},
  pages={417--421},
  year={2023},
  doi={10.1145/3594536.3595170}
}

@article{rabelo2022overview,
  title={Overview and discussion of the competition on legal information extraction/entailment (COLIEE) 2021},
  author={Rabelo, Juliano and Goebel, Randy and Kim, Mi-Young and Kano, Yoshinobu and Yoshioka, Masaharu and Satoh, Ken},
  journal={The Review of Socionetwork Strategies},
  volume={16},
  number={1},
  pages={111--133},
  year={2022},
  publisher={Springer},
  doi={10.1007/s12626-022-00105-z}
}

@book{gwet2014handbook,
  author    = {Gwet, Kilem L.},
  title     = {Handbook of Inter-Rater Reliability: The Definitive Guide
               to Measuring the Extent of Agreement Among Raters},
  edition   = {4},
  year      = {2014},
  publisher = {Advanced Analytics, LLC},
  address   = {Gaithersburg, MD},
  isbn      = {978-0-9708062-8-4}
}

@article{feinstein1990high,
  author  = {Feinstein, Alvan R. and Cicchetti, Domenic V.},
  title   = {High agreement but low kappa: {I}. The problems of two paradoxes},
  journal = {Journal of Clinical Epidemiology},
  volume  = {43},
  number  = {6},
  pages   = {543--549},
  year    = {1990},
  doi     = {10.1016/0895-4356(90)90158-L}
}

@article{wongpakaran2013comparison,
  author  = {Wongpakaran, Nahathai and Wongpakaran, Tinakon and Wedding, Danny
             and Gwet, Kilem L.},
  title   = {A comparison of {C}ohen's Kappa and {G}wet's {AC1} when
             calculating inter-rater reliability coefficients: a study
             conducted with personality disorder samples},
  journal = {BMC Medical Research Methodology},
  number = {1},
  volume  = {13},
  pages   = {61},
  year    = {2013},
  publisher = {Springer},
  doi     = {10.1186/1471-2288-13-61}
}

@article{coupette2021measuring,
  title={Measuring law over time: a network analytical framework with an application to statutes and regulations in the United States and Germany},
  author={Coupette, Corinna and Beckedorf, Janis and Hartung, Dirk and Bommarito, Michael and Katz, Daniel Martin},
  journal={Frontiers in Physics},
  volume={9},
  pages={658463},
  year={2021},
  publisher={Frontiers Media SA},
  doi={10.3389/fphy.2021.658463}
}

@article{ovadek2021analysing,
  title={Analysing eu treaty-making and litigation with network analysis and natural language processing},
  author={Ov{\'a}dek, Michal and Dyevre, Arthur and Wigard, Kyra},
  journal={Frontiers in Physics},
  volume={9},
  pages={657607},
  year={2021},
  publisher={Frontiers Media SA},
  doi={10.3389/fphy.2021.657607}
}

@book{digitpa2012linee,
  title={Linee guida per la marcatura dei documenti normativi secondo gli standard Normeinrete},
  author={Giovannini, Maria Pia and Palmirani, Monica and Francesconi, Enrico},
  year={2012},
  publisher={European Press Academic Publishing Firenze},
  isbn={978-88-8398-076-3}
}

@inproceedings{holzenberger2023connecting,
  title={Connecting symbolic statutory reasoning with legal information extraction},
  author={Holzenberger, Nils and Van Durme, Benjamin},
  booktitle={Proceedings of the Natural Legal Language Processing Workshop 2023},
  pages={113--131},
  year={2023},
  doi={10.18653/v1/2023.nllp-1.12},
  note={This entry previously appeared twice in the bibliography (duplicate key); the duplicate has been removed}
}

@inproceedings{dahan2023openjustice,
  title={Openjustice. ai: A global open-source legal language model},
  author={Dahan, Samuel and Bhambhoria, Rohan and Liang, David and Zhu, Xiaodan},
  booktitle={Legal Knowledge and Information Systems: JURIX 2023: The Thirty-sixth Annual Conference, Maastricht, the Netherlands, 18--20 December 2023},
  pages={387--390},
  year={2023},
  organization={IOS Press},
  doi={10.3233/FAIA230995}
}

@incollection{bench2024computational,
  title={Computational models of legal argument},
  author={Bench-Capon, TJM and Atkinson, Katie and Bex, Floris and Prakken, Henry and Verheij, HB},
  booktitle={Handbook of Formal Argumentation},
  pages={1--123},
  year={2024},
  publisher={College Publications},
  note={No DOI available; College Publications does not assign DOIs to this handbook}
}

@book{ashley2017artificial,
  title={Artificial intelligence and legal analytics: new tools for law practice in the digital age},
  author={Ashley, Kevin D},
  year={2017},
  publisher={Cambridge University Press},
  doi={10.1017/9781316761380}
}

@book{maccormick1994legal,
  title={Legal reasoning and legal theory},
  author={MacCormick, Neil},
  year={1994},
  publisher={Clarendon Press},
  isbn={978-0-19-876384-0}
}

@article{dahl2024large,
  title={Large legal fictions: Profiling legal hallucinations in large language models},
  author={Dahl, Matthew and Magesh, Varun and Suzgun, Mirac and Ho, Daniel E},
  journal={Journal of Legal Analysis},
  volume={16},
  number={1},
  pages={64--93},
  year={2024},
  publisher={Oxford University Press UK},
  doi={10.1093/jla/laae003}
}

@article{brin1998anatomy,
  title={The anatomy of a large-scale hypertextual web search engine},
  author={Brin, Sergey and Page, Lawrence},
  journal={Computer networks and ISDN systems},
  volume={30},
  number={1-7},
  pages={107--117},
  year={1998},
  publisher={Elsevier},
  doi={10.1016/S0169-7552(98)00110-X}
}

@incollection{bhambhoria2024evaluating,
  title={Evaluating AI for Law: Bridging the Gap with Open-Source Solutions},
  author={Bhambhoria, Rohan and Dahan, Samuel and Li, Jonathan and Zhu, Xiaodan},
  booktitle={Compliance for Artificial Intelligence Systems},
  series={Lecture Notes in Computer Science},
  volume={14377},
  pages={59--74},
  year={2026},
  publisher={Springer},
  doi={10.1007/978-3-032-12795-2_5},
  note={Preprint available as arXiv:2404.12349 (2024)}
}

@article{liu2024deepseek,
  title={Deepseek-v3 technical report},
  author={Liu, Aixin and Feng, Bei and Xue, Bing and Wang, Bingxuan and Wu, Bochao and Lu, Chengda and Zhao, Chenggang and Deng, Chengqi and Zhang, Chenyu and Ruan, Chong and others},
  journal={arXiv preprint arXiv:2412.19437},
  year={2024}
}

@book{villa2023graduazione,
  title={La graduazione delle questioni di merito. Ammissibilit{\`a} e profili dinamici},
  author={Villa, Alberto},
  volume={149},
  year={2023},
  publisher={Giappichelli}
}

@book{liebman2007manuale,
  title={Manuale di diritto processuale civile. Principi},
  author={Liebman, Enrico Tullio and Colesanti, Vittorio and Merlin, Elena and Ricci, Edoardo F},
  year={2007},
  publisher={Giuffr{\`e} Editore}
}

@incollection{Ricci1987Accertamento,
  title={Accertamento giudiziale},
  author={Ricci, Francesco},
  booktitle={digesto discipline privatistiche},
  year={1987},
  publisher={utet}
}

@inproceedings{bench2020explaining,
  title={Explaining legal decisions using IRAC},
  author={Bench-Capon, Trevor},
  booktitle={CEUR Workshop Proceedings},
  volume={2669},
  pages={74--83},
  year={2020},
  organization={CEUR-WS. org}
}

@incollection{ruf2024aristotle,
  title={Aristotle is Long Dead But His Wisdom Rules Us from the Grave: A Comparison of the Application of Logic in Legal Reasoning in Common Law and Civil Law Systems},
  author={Ruf, Anna and Yin, Kenneth},
  booktitle={Comparative Law: Unraveling Global Legal Systems},
  pages={1--23},
  year={2024},
  publisher={Springer}
}

\appendix
\section{Two-step prompt}
In this appendix we report the two-step prompts used for the extraction. As explained in section \ref{sec:LLM_extraction}, prompt \ref{lst:prompt} was used for shorter judgments while a two-step approach was used for longer ones. 

\begin{lstlisting}[caption={First part of the two-step prompt (translated in English). The goal of this prompt is to extract and summarize all issues within the judgment}, label={lst:prompt_multistep_questioni}]
Inside the curly braces there is an Italian tax court decision. Your task is to identify and extract the main legal issues (generally one or two) that the judge addresses and resolves in order to decide the dispute. Consider only the issues on which the judge rules explicitly and with reasoning.

For each issue, perform the following:
- Formulate the legal issue clearly, as a self-contained and autonomous question. The formulation must follow the structure "Whether + clause" and must be understandable on its own, without referring to the full decision or other issues. Be concise.
- Indicate the outcome of the issue, i.e., the answer to the legal question. Include only the resolution of the issue without mentioning other rulings of the judge.
- Write an abstractive summary of approximately 100 words, including:
  -- the facts relevant to the issue
  -- the main legislative and case law references relevant to the issue
  -- a synthesis of the judge's reasoning in deciding the issue
  -- the conclusion reached by the judge

The summary must be self-contained and understandable without reference to the full decision.

Always use precise legal language. Include only the information relevant to each issue.

XML output format:
<issues>
  <issue title="issue title">
    <text>text of the legal issue (~30 words)</text>
    <issue_outcome in_favor_of="tax authority|taxpayer|other">testo esito</issue_outcome>
    <summary>summary (~100 words) of the issue</summary>
  </issue>
</issues>

Output ONLY the XML block and nothing else.
\end{lstlisting}

\begin{lstlisting}[caption={Second part of the two-step prompt (translated in English). The text of the judgment and the <text> field of one of the issues extracted in the previous step are passed to the LLM along with the prompt.}, label={lst:prompt_multistep_reasoning}]
Inside curly brackets is an Italian tax judgment. Inside square brackets, a legal issue already identified in the same judgment is also provided. Your task is to extract the judge's reasoning from the decision with reference to that issue. Structure the reasoning as follows:

<factual_premises>: list the facts, actions, events, or data of the case considered relevant to the issue. Each fact must be described with sufficient precision to allow another judge to decide a similar issue consistently.
<legal_references>: list exhaustively ALL AND ONLY the legal references (statutory provisions, case law, general principles, administrative practice) used by the judge to resolve the issue, as they appear in the judgment. Order them by decreasing importance.
<citation_reason>: for only those legal references discussed in greater detail, explain why the judge cites them. Consider only references that are discussed in depth in the judgment.
<judge_reasoning>: describe in free text how the judge applied the legal references to the specific case, leading to the decision. Do not include information already stated in the factual premises. Use approximately fewer than 120 words.

Always use precise legal language.

XML output format:
<factual_premises>
  <item>factual premise 1</item>
  <item>factual premise 2</item>
</factual_premises>
<legal_references>
  <item id="D1" type="caselaw" ref="Supreme Court no. nnnn/yyyy"/>
  <item id="D2" type="leg" ref="Art. xx, para. z, Legislative Decree no. nnn/yyyy"/>
  <item id="D3" type="caselaw" ref="Supreme Court no. nnnn/yyyy, Supreme Court no. mmmm/yyyy"/>
</legal_references>
<citation_reason>
  <item ref_id="D1">reason for citing legal reference</item>
  <item ref_id="D2">reason for citing legal reference</item>
</citation_reason>
<judge_reasoning>reasoning text (fewer than 150 words)</judge_reasoning>

In the ref field of legal references, always indicate the exact citation details of the source. Use the type field to classify the reference:
type="caselaw" for case law. If multiple decisions are cited consecutively, group them in a single item;
type="leg" for legislation;
type="princ" for general principles of law (in this case, ref contains only the name of the principle);
type="admin_pract" for administrative practice.

Include only information relevant to the given issue.
Generate only the XML block and nothing else.
\end{lstlisting}
\section{Issue text correction}

\begin{lstlisting}[caption={Few-shot prompt used to correct the <text> field.}, label={lst:issue_text_correction}]
You are an expert in the Italian language and tax law. Your task is to correct legal issue questions extracted from Italian tax court decisions.

CORRECTION RULES:
1. Correct structure: The questions must be indirect interrogatives starting with "Whether" followed by a subjunctive verb, expressing the legal issue. The initial "Whether" must be followed by the subject of the sentence. They must not be conditional clauses introduced by "if/whether" in a hypothetical sense.
2. Indirect interrogative form: "Whether" introduces the main clause (e.g. "Whether the taxpayer is required...", "Whether there exists..."), not a subordinate conditional clause. Do not use a final question mark.
3. Agreement and syntax: Correct agreement errors and add missing relative pronouns.

EXAMPLES OF CORRECTION:

INCORRECT:
Whether a taxpayer produces waste disposed of independently through a private company, without the Municipality having provided any collection service, is exempt from the payment of the waste tax.

CORRECT:
Whether a taxpayer who produces waste disposed of independently through a private company, without the Municipality having provided any collection service, is exempt from the payment of the waste tax.

INCORRECT:
Whether a business displays a sign identifying the premises of the activity, there is exemption from the municipal advertising tax.

CORRECT:
Whether there is an exemption from the municipal advertising tax for a business that displays a sign identifying the premises of the activity.

INCORRECT:
The Tax Authority, not having appeared in the first-instance proceedings, has the necessary standing to appeal the first-instance judgment?

CORRECT:
Whether the Tax Authority, not having appeared in the first-instance proceedings, has the necessary standing to appeal the first-instance judgment.

INCORRECT:
Whether an asset is subject to judicial seizure, whether the taxpayer is required to pay IMU despite the seizure and lack of availability of the asset.

CORRECT:
Whether, in the case where an asset is subject to judicial seizure, the taxpayer is required to pay IMU despite the seizure and lack of availability of the asset.

INSTRUCTIONS:
IMPORTANT: Only about 15% of questions contain errors. First, assess whether the question actually needs correction. If it does not, respond only: QUESTION ALREADY CORRECT

Correct the question ONLY if it presents one or more of the following issues:
- Direct interrogative form ("?") instead of indirect form
- Agreement or syntactic errors
- Confused structure (e.g. double "Whether", conditional periods instead of indirect interrogatives)
- Missing necessary relative pronouns

Output ONLY the corrected question.
If the question is already correct, write: QUESTION ALREADY CORRECT

QUESTION TO EVALUATE:
\end{lstlisting}

\section{Prompts in Italian}
\label{app:prompts_ita}
In this section we report all prompts in their original form, that is, in Italian. All of the documents involved in our analysis are in fact in Italian, including judgments, prompts, and XML outputs.

\begin{lstlisting}[caption={Single step prompt}, label={lst:prompt_ita}]
All'interno delle parentesi graffe è riportata una sentenza tributaria italiana.
Il tuo compito è identificare ed estrarre le principali questioni giuridiche (una o massimo due) che il giudice affronta e risolve per decidere la controversia. Considera esclusivamente le questioni su cui il giudice si pronuncia in modo esplicito e motivato.
Per ciascuna questione, svolgi le seguenti operazioni:
- Formula la questione giuridica in modo chiaro, autosufficiente e autonomo. La formulazione deve essere nella forma "Se + congiuntivo" e deve essere comprensibile da sola, senza dover consultare l'intera sentenza o altre questioni.
- Indica l'esito della questione, cioè la risposta al quesito giuridico. 
- Scomponi il ragionamento del giudice nelle seguenti parti: 
-- <premesse_fatto>: elenca le circostanze, azioni, eventi o dati del caso ritenuti rilevanti per la questione.
-- <lista_riferimenti_diritto>: elenca in modo completo TUTTI i riferimenti giuridici (norme, giurisprudenza, principi generali, prassi amministrativa) rilevanti per la questione, così come appaiono nella sentenza. Ordinali per importanza decrescente.
-- <motivo_citazione>: per i soli riferimenti giuridici trattati più in dettaglio spiega perché il giudice li cita. Considera solo i riferimenti discussi in modo argomentato e approfondito nella sentenza.
-- <ragionamento_giudice>: descrivi in testo libero come il giudice ha applicato i riferimenti giuridici al caso concreto, fino a arrivare alla decisione. Non includere le informazioni già presenti nelle premesse di fatto.
- Redigi un riassunto astrattivo della questione, tra 100 e 150 parole, che includa: fatto, diritto, ragionamento e conclusione. Il riassunto deve essere autosufficiente e completo, comprensibile senza fare riferimento alla sentenza completa o ad altre questioni.
Requisiti generali:
- Ogni questione deve essere autonoma, autosufficiente e formulata in linguaggio giuridico preciso e dettagliato.
- Il fatto deve essere descritto con sufficiente precisione da permettere a un altro giudice di decidere una questione analoga in modo coerente.


Formato di output richiesto: esclusivamente in XML, come nello schema seguente:
<questioni>
  <questione title="titolo questione">
    <text>testo della questione giuridica</text>
    <esito_questione favorevole_a="ufficio|contribuente|other">testo esito</esito_questione>
    <premesse_fatto>
      <item>premessa di fatto 1</item>
      <item>premessa di fatto 2</item>
    </premesse_fatto>
    <lista_riferimenti_diritto>
      <item id="D1" type="jur" ref="Cassazione n. nnnn/yyyy"/>
      <item id="D2" type="norm" ref="art. xx comma z d.lgs n. nnn/yyyy"/>
    </lista_riferimenti_diritto>
    <motivo_citazione>
      <item id_ref="D1"> motivo citazione diritto </item>
      <item id_ref="D2"> motivo citazione diritto </item>
    </motivo_citazione>
    <ragionamento_giudice>testo ragionamento (~100 parole)</ragionamento_giudice>
    <riassunto>testo del riassunto (~100 parole)</riassunto>
  </questione>
</questioni>
Nel campo ref dei riferimenti giuridici indica sempre gli estremi esatti dell'atto. Usa il campo type per classificare il riferimento:
- type="jur" per giurisprudenza. Se più sentenze vengono citate consecutivamente, raggruppale in un unico item;
- type="norm" per norme di legge;
- type="princ" per principi generali del diritto (in questo caso, ref contiene esclusivamente il nome del principio);
- type="prassi_amm" per la prassi amministrativa.
Genera solo il blocco XML e nulla oltre.

\end{lstlisting}

\begin{lstlisting}[caption={First prompt of the two-step approach. This prompt extracts issues together with their respective outcomes and summaries}, label={lst:prompt_multistep_issues_ita}]
All'interno delle parentesi graffe è riportata una sentenza tributaria italiana. Il tuo compito è identificare ed estrarre le principali questioni giuridiche (generalmente una o due) che il giudice affronta e risolve per decidere la controversia. Considera esclusivamente le questioni su cui il giudice si pronuncia in modo esplicito e motivato.
Per ciascuna questione, svolgi le seguenti operazioni:
- Formula la questione giuridica in modo chiaro, autosufficiente e autonomo. La formulazione deve essere nella forma "Se + congiuntivo" e deve essere comprensibile da sola, senza dover consultare l'intera sentenza o altre questioni. Sii sintetico.
- Indica l'esito della questione, cioè la risposta al quesito giuridico. Includi solo la soluzione della questione senza indicare altre disposizioni del giudice.
- Scrivi un riassunto astrattivo di circa 100 parole della questione, che includa:
-- i fatti rilevanti per la questione
-- i riferimenti normativi e giurisprudenziali principali rilevanti per la questione
-- la sintesi del ragionamento del giudice per decidere la questione
-- la conclusione della questione raggiunta dal giudice
Il riassunto deve essere autonomo e comprensibile senza riferimento alla sentenza completa.

Usa sempre un linguaggio giuridico preciso. Includi solo le informazioni rilevanti per ciascuna questione.

Formato di output XML:
<questioni>
  <questione title="titolo questione">
    <text>testo della questione giuridica (~30 parole)</text>
    <esito_questione favorevole_a="ufficio|contribuente|other">testo esito</esito_questione>
    <riassunto> riassunto (~100 parole) della questione</riassunto>
  </questione>
</questioni>
 Genera solo il blocco XML e nulla oltre.

\end{lstlisting}

\begin{lstlisting}[caption={Second prompt of the two-step approach. This prompt extracts the reasoning (facts, legal references, citation reasons, judge reasoning).}, label={lst:prompt_multistep_reasoning_ita}]
All'interno delle parentesi graffe è riportata una sentenza tributaria italiana. Tra parentesi quadre è inoltre fornita una questione giuridica già identificata all'interno della stessa sentenza. Il tuo compito è estrarre il ragionamento del giudice dalla sentenza in riferimento alla questione. Suddividi il ragionamento come segue:
- <premesse_fatto>: elenca le circostanze, azioni, eventi o dati del caso ritenuti rilevanti per la questione. Il fatto deve essere descritto con sufficiente precisione da permettere a un altro giudice di decidere una questione analoga in modo coerente.
- <lista_riferimenti_diritto>: elenca in modo completo TUTTI E SOLI i riferimenti giuridici (norme, giurisprudenza, principi generali, prassi amministrativa) usati dal giudice per risolvere la questione data, così come appaiono nella sentenza. Ordinali per importanza decrescente.
- <motivo_citazione>: per i soli riferimenti giuridici trattati più in dettaglio spiega perché il giudice li cita. Considera solo i riferimenti discussi in modo argomentato e approfondito nella sentenza.
- <ragionamento_giudice>: descrivi in testo libero come il giudice ha applicato i riferimenti giuridici al caso concreto, fino a arrivare alla decisione. Non includere le informazioni già presenti nelle premesse di fatto. Usa meno di 120 parole circa.
Usa sempre un linguaggio giuridico preciso.
Formato di output XML:
<premesse_fatto>
  <item>premessa di fatto 1</item>
  <item>premessa di fatto 2</item>
</premesse_fatto>
<lista_riferimenti_diritto>
  <item id="D1" type="jur" ref="Cassazione n. nnnn/yyyy"/>
  <item id="D2" type="norm" ref="art. xx comma z dlgs n. nnn/yyyy"/>
  <item id="D3" type="jur" ref="Cassazione n. nnnn/yyyy, Cassazione n. mmmm/yyyy"/>
</lista_riferimenti_diritto>
<motivo_citazione>
  <item id_ref="D1"> motivo citazione diritto </item>
  <item id_ref="D2"> motivo citazione diritto </item>
</motivo_citazione>
<ragionamento_giudice>testo ragionamento (meno di 150 parole)</ragionamento_giudice>
Nel campo ref dei riferimenti giuridici indica sempre gli estremi esatti dell'atto. Usa il campo type per classificare il riferimento:
- type="jur" per giurisprudenza. Se più sentenze vengono citate consecutivamente, raggruppale in un unico item;
- type="norm" per norme di legge;
- type="princ" per principi generali del diritto (in questo caso, ref contiene esclusivamente il nome del principio);
- type="prassi_amm" per la prassi amministrativa.
Includi solo le informazioni rilevanti per la questione data.
Genera solo il blocco XML e nulla oltre.

\end{lstlisting}

\begin{lstlisting}[caption={Few-shot prompt used to correct the <text> field.}, label={lst:issue_text_correction_ita}]
Sei un esperto di lingua italiana e diritto tributario. Il tuo compito è correggere quesiti di questioni giuridiche estratte da sentenze tributarie italiane.

REGOLE DI CORREZIONE:
1. Struttura corretta: I quesiti devono essere interrogative indirette che iniziano con "Se" + congiuntivo per porre la questione giuridica. Il "Se" iniziale deve essere seguito dal soggetto della frase. Non devono essere periodi ipotetici con "Se" che introduce una condizione.
2. Forma interrogativa indiretta: "Se" introduce la proposizione principale (es. "Se il contribuente sia tenuto...", "Se sussista..."), non una subordinata ipotetica. Non usare il punto interrogativo finale.
4. Concordanza e sintassi: Correggi errori di concordanza, aggiungi pronomi relativi mancanti.
ESEMPI DI CORREZIONE:

SCORRETTO:
Se un contribuente produca rifiuti smaltiti autonomamente tramite ditta privata, senza che il Comune abbia svolto alcun servizio di raccolta, sia esente dal pagamento della TARI.
CORRETTO:
Se un contribuente che produce rifiuti smaltiti autonomamente tramite ditta privata, senza che il Comune abbia svolto alcun servizio di raccolta, sia esente dal pagamento della TARI.

SCORRETTO:
Se un'impresa espone un'insegna di esercizio che contraddistingue la sede dell'attività, sussista l'esenzione dall'imposta comunale sulla pubblicità.
CORRETTO:
Se sussista l'esenzione dall'imposta comunale sulla pubblicità per un'impresa che espone un'insegna di esercizio che contraddistingue la sede dell'attività.

SCORRETTO:
L'Agenzia delle Entrate, non essendosi costituita nel giudizio di primo grado, ha la legittimazione necessaria per proporre appello avverso la sentenza di primo grado?
CORRETTO:
Se l'Agenzia delle Entrate, non essendosi costituita nel giudizio di primo grado, abbia la legittimazione necessaria per proporre appello avverso la sentenza di primo grado.

SCORRETTO:
Se un immobile sia soggetto a sequestro giudiziario, se il contribuente sia tenuto al pagamento dell'IMU nonostante il sequestro e la mancata disponibilità del bene.
CORRETTO:
Se, nel caso in cui un immobile sia soggetto a sequestro giudiziario, il contribuente sia tenuto al pagamento dell'IMU nonostante il sequestro e la mancata disponibilità del bene.

ISTRUZIONI:
IMPORTANTE: Solo il 15% circa dei quesiti presenta errori. Prima di tutto, valuta se il quesito necessita davvero di correzioni. In caso contrario scrivi unicamente: QUESITO GIÀ CORRETTO

Correggi il quesito SOLO se presenta uno o più di questi problemi:
- Forma interrogativa diretta (con "?") invece che indiretta
- Errori di concordanza o sintassi
- Struttura confusa (es. doppi "Se", periodi ipotetici invece di interrogative indirette)  
- Mancanza di pronomi relativi necessari


In output produci ESCLUSIVAMENTE il quesito corretto
Se il quesito è già corretto, scrivi: QUESITO GIÀ CORRETTO
QUESITO DA VALUTARE:
\end{lstlisting}

\section{Extraction results averaged per judgment and per issue}
\label{app:macro}

The main text reports precision, recall, and F1 computed by \emph{pooling} all
items (issues or citations) across the relevant judgments before computing the
metric. While this is a natural statistic, as it gives information about how accurate a single issue or citation is, it can be skewed by the judgments that contain many issues, or the issues with many associated references.
Here we report, alongside the pooled figures, the corresponding scores
obtained by computing the metric separately on each unit (each judgment for
issues, each present issue for citations) and then averaging, so that every unit
contributes equally regardless of how many items it contains. Each table places
the two views side by side.

\subsection{Issue extraction}

Averaging per judgment raises every issue score relative to pooling
(Tables~\ref{tab:issue_pairwise_macro} and~\ref{tab:issue_n35_macro}). First, we remark that the inter annotator agreement improves. The LLM's precision and recall also increase significantly (recall goes from $69.9\%$ to $86.3\%$ and from $82.3\%$ to $92.4\%$) when averaged per-judgment. This confirms that the model performs well on typical judgments and is penalized mainly on a few harder ones.

\begin{table}[!htbp]
\centering
\caption{Pairwise agreement on issue extraction ($N{=}20$ shared judgments):
pooled vs.\ per-judgment averaged. Each row treats the second system as ground
truth.}
\label{tab:issue_pairwise_macro}
\begin{tabular}{l cc cc}
\toprule
 & \multicolumn{2}{c}{Pooled} & \multicolumn{2}{c}{Per-judgment} \\
\cmidrule(lr){2-3}\cmidrule(lr){4-5}
Comparison & P & R  & P & R \\
\midrule
$A_1 \mid A_2$  & 78.7  & 100.0 & 91.3  & 100.0  \\
$A_2 \mid A_1$  & 100.0 & 78.7  & 100.0 & 91.3   \\
LLM $\mid A_1$  & 93.3  & 59.6  &  95.0  & 82.7   \\
LLM $\mid A_2$  & 93.3  & 75.7  &  95.0  & 90.0   \\
\bottomrule
\end{tabular}
\end{table}

\begin{table}[!htbp]
\centering
\caption{LLM issue extraction against each annotator's full set ($N{=}35$):
pooled vs.\ per-judgment averaged.}
\label{tab:issue_n35_macro}
\begin{tabular}{l cc cc}
\toprule
 & \multicolumn{2}{c}{Pooled} & \multicolumn{2}{c}{Per-judgment} \\
\cmidrule(lr){2-3}\cmidrule(lr){4-5}
 & LLM $\mid A_1$ & LLM $\mid A_2$ & LLM $\mid A_1$ & LLM $\mid A_2$ \\
\midrule
Precision & 94.4 & 94.4 & 95.7 & 95.7 \\
Recall    & 69.9 & 82.3 & 86.3 & 92.4 \\
\bottomrule
\end{tabular}
\end{table}

\subsection{Citation extraction}

For citations the averaging unit is the present issue, and the effects are similar (Tables~\ref{tab:cit_pairwise_macro} and~\ref{tab:cit_n35_macro}). Once again, the precision and recall show some improvement in the per-issue case.
\begin{table}[!htbp]
\centering
\caption{Pairwise agreement on citation extraction ($N{=}20$, $28$~present
issues): pooled vs.\ per-issue averaged.}
\label{tab:cit_pairwise_macro}
\begin{tabular}{l cc cc}
\toprule
 & \multicolumn{2}{c}{Pooled} & \multicolumn{2}{c}{Per-issue} \\
\cmidrule(lr){2-3}\cmidrule(lr){4-5}
Comparison & P & R & P & R  \\
\midrule
$A_1 \mid A_2$ & 98.5 & 95.7  & 99.5 & 98.3  \\
$A_2 \mid A_1$ & 95.7 & 98.5  & 98.3 & 99.5  \\
LLM $\mid A_1$ & 72.2 & 77.6  & 72.4 & 80.1  \\
LLM $\mid A_2$ & 72.2 & 75.4  & 72.4 & 78.8  \\
\bottomrule
\end{tabular}
\end{table}

\begin{table}[!htbp]
\centering
\caption{LLM citation extraction against each annotator's full set ($N{=}35$):
pooled vs.\ per-issue averaged.}
\label{tab:cit_n35_macro}
\begin{tabular}{l cc cc}
\toprule
 & \multicolumn{2}{c}{Pooled} & \multicolumn{2}{c}{Per-issue} \\
\cmidrule(lr){2-3}\cmidrule(lr){4-5}
 & LLM $\mid A_1$ & LLM $\mid A_2$ & LLM $\mid A_1$ & LLM $\mid A_2$ \\
\midrule
Precision & 72.7 & 73.9 & 74.9 & 75.9 \\
Recall    & 70.3 & 80.6 & 85.6 & 83.8 \\
\bottomrule
\end{tabular}
\end{table}

\subsection{Union and intersection ground truth ($N{=}50$)}
\label{app:union_inter}

The main text reports LLM performance separately against each annotator. An
alternative is to combine the two annotators into a single ground truth over all
$N{=}50$ judgments. For the $30$~judgments evaluated by a single annotator, the
ground truth is that annotator's set. For the $20$~shared judgments, we consider
two strategies:
\begin{itemize}
    \item \textbf{Union}: an item (issue or citation) is in the ground truth if
    \emph{either} annotator identified it. This produces a larger reference set
    and thus a more demanding recall target.
    \item \textbf{Intersection}: an item is in the ground truth only if
    \emph{both} annotators identified it. This produces a stricter reference set
    that may increase measured recall but can also affect precision, since an
    LLM-extracted item confirmed by only one annotator counts as a false
    positive.
\end{itemize}

\paragraph{Issue extraction.}
Table~\ref{tab:issue_union_inter} reports the results under both definitions.
Precision is identical ($94.9\%$) because both annotators agree on which LLM
issues are present in every shared judgment. The difference appears in recall:
the union ground truth includes more annotator issues (lowering recall to
$75.5\%$), while the intersection ground truth is smaller (recall rises to
$84.1\%$).

\begin{table}[!htbp]
\centering
\caption{LLM issue extraction with union/intersection ground truth ($N{=}50$).}
\label{tab:issue_union_inter}
\begin{tabular}{l cc}
\toprule
 & Union & Intersection \\
\midrule
Precision (pooled)       & 94.9 & 94.9 \\
Recall (pooled)          & 75.5 & 84.1 \\
Precision (per-judgment) & 96.0 & 96.0 \\
Recall (per-judgment)    & 89.1 & 92.0 \\
\bottomrule
\end{tabular}
\end{table}

\paragraph{Citation extraction.}
Table~\ref{tab:cit_union_inter} reports the analogous results for citations. Here
the precision values differ slightly between union and intersection, because the
number of LLM citations matching the ground truth changes depending on the
definition.\footnote{For $3$ of $28$ shared issues, the exact count of LLM
citations falling in the intersection of both annotators' lists
($|\text{LLM} \cap A_1 \cap A_2|$) cannot be determined from the available data;
we use the midpoint of tight inclusion-exclusion bounds, with a maximum error of
${\pm}1$ citation per affected issue.}

\begin{table}[!htbp]
\centering
\caption{LLM citation extraction with union/intersection ground truth ($N{=}50$,
$74$~present issues).}
\label{tab:cit_union_inter}
\begin{tabular}{l cc}
\toprule
 & Union & Intersection \\
\midrule
Precision (pooled)    & 74.3 & 73.0 \\
Recall (pooled)       & 75.0 & 75.0 \\
Precision (per-issue) & 77.0 & 75.9 \\
Recall (per-issue)    & 87.1 & 85.9 \\
\bottomrule
\end{tabular}
\end{table}
\section{Annotators scoring criteria}
\label{app:scoring_criteria}

In this appendix we present the guidelines followed by the annotators.
As a general rule, all fields extracted by the LLM must be validated. For each field, both quantitative and qualitative metrics are defined. Quantitative metrics are numerically measurable, although they may still involve a certain degree of expert judgment (i.e. number of extracted legal issues, number of extracted relevant citations). Qualitative metrics rely on the expert’s assessment, typically expressed as a numerical score corresponding to predefined, albeit arbitrary, thresholds.

\textbf{Independent issue identification}
The LLM-generated output to be evaluated is the one provided in the section \verb|<text>|
A legal issue is described as a legal question explicitly decided by the judge in the reasoning section of the judgment, requiring the interpretation or application of a legal source, and supported by argumentative content sufficient to fill the structured fields of the XML format.
The annotator shall manually extract from the argumentative part of the judgement all the relevant legal issues according to the previous definition.

\textbf{Issue alignment and scoring}

The annotator should mark the presence of each LLM-extracted issue as:
\begin{itemize}
    \item TRUE if it matches an issue in the annotator’s list;
    \item FALSE if it does not correspond to any issue the annotator considered present in the judgment.
\end{itemize}
The annotator shall indicate the number of legal issues present in its list but not extracted by the LLM (the minimum number of legal issues which, in addition to those already extracted, would be required to cover the entire content of the judgment).
Any further step of the assessment shall concern only the legal issues present in the judgement.
For each LLM-extracted legal issue, the annotator shall assess:
\begin{itemize}
    \item generality: whether the automatically extracted legal issue is neither overly specific nor excessively general; the optimal level is the one that a legal expert would expect to find among the results of a search and that most closely approximates the correspondent manually extracted legal issue;
    \item legal language: whether legal terminology and the wording are appropriate; 
\end{itemize}
The assessment is carried out by assigning a score ranging from 1 (completely inappropriate) to 5 (optimal level) for both generality and legal language of each extracted issue.

\paragraph{Citation}\phantom{.}
\\
\\
\textit{Independent citation identification }
The LLM-generated output to be evaluated is the one provided in the section \verb|<legal_references>|.
Given the text of the legal issue extracted by the LLM, the annotator shall compile a list of the legal references contained in the portion of the judgment correspondent to the legal issue under evaluation. The annotator shall select only the references that are used by the judge to solve the question. That means that not all the legal references mentioned in the judgment shall be considered, but only those relevant.
Citations shall be listed in the order in which they appear in the judgment.
That list is meant to be compiled without looking at \verb|<legal_references>| section, which is the list of relevant citations extracted by the LLM.

\textit{Citation reason evaluation}
The LLM-generated output to be evaluated is the one provided in the section \verb|<citation_reason>|
Since not all the references quoted are relevant, the annotator shall also evaluate the motivation given by the LLM for their extraction. Normally, not all the extracted references are motivated, the annotator shall assess the motivations actually present. No assessment is required of the criteria the LLM used to select the reference to be explained.
Citation reason is assessed by the sum of citation rationale scores: for each rationale, the annotator assigns a score of 1 if the rationale is both contained in the judgment and faithfully represents its content, and 0 otherwise (e.g. if the LLM presented the rationale for three out of five citation extracted, the annotator shall assign a maximum score of 3 and a minimum of 0)

\paragraph{Summary}
The LLM-generated output to be evaluated is the one provided in the section \verb|<summary>|
Assessment of the summary is carried out by assessing a score ranging from 1 (completely inappropriate) to 5 (optimal level) to each of the following aspects:
\begin{itemize}
    \item Correctness as the absence of elements that are not present in the source document;
    \item Form as coherence, readability, syntactic and grammatical correctness, and adherence to legal terminology;
    \item Completeness as the inclusion of all relevant elements of the decision (key points, legal issues, and essential factual information) and the comprehensive representation of the source’s content;
    \item Satisfaction as the degree of satisfaction with the global quality of the summary.
\end{itemize}

\paragraph{Fact}
The LLM-generated output to be evaluated is the one provided in the section \verb|<factual_premises>|.
The annotator shall provide an overall assessment of the quality of the fact as summarised by the LLM. The overall assessment is carried out by assessing a comprehensive score ranging from 1 (completely inappropriate) to 5 (optimal level).
In assessing the overall satisfaction the annotator shall take into account:
\begin{itemize}
    \item the absence of factual elements that are not supported by the source document;
    \item the coherence, readability, syntactic and grammatical accuracy, and appropriate use of legal terminology;
    \item the inclusion of all elements of the decision that are relevant to the extracted content.
\end{itemize}

\paragraph{Reasoning}
The LLM-generated output to be evaluated is the one provided in the section \verb|<judge_reasoning>|
The annotator shall provide an overall assessment of the quality of the reasoning of the judge as summarised by the LLM. The general assessment is carried out by assessing a comprehensive satisfaction score ranging from 1 (completely inappropriate) to 5 (optimal level).
In assessing the overall satisfaction the annotator shall take into account:
\begin{itemize}
    \item absence of elements that are not present in the source document (e.g. the statement made by the judge in the decision is actually motivated based on those reasons);
    \item correct use of logical connections (e.g. the reasoning steps extracted by the LLM actually bring to the conclusion);
    \item coherence, readability, syntactic and grammatical correctness, and adherence to legal terminology;
    \item inclusion of all relevant elements of the decision and the comprehensive representation of the source’s content.
\end{itemize}

\paragraph{Outcome}
The LLM-generated output to be evaluated is the one provided in the section \verb|<issue_outcome>|.
The annotator shall provide an overall assessment of the outcome of the procedure as summarised by the LLM. The general assessment is carried out by assigning a satisfaction score ranging from 1 (completely inappropriate) to 5 (optimal level). It refers to the degree of satisfaction with the global quality of the outcome formulation.

\section{Cohen's $\kappa$ for the Likert dimensions}
\label{app:kappa}

For completeness, Table~\ref{tab:quality_scores_kappa} reports Cohen's $\kappa$ with linear and quadratic weights and the Pearson correlation $r$ for the same dimensions as Table~\ref{tab:quality_scores}. These chance-corrected statistics paint a substantially different picture from the saturation-robust statistics reported in the main text. On the two dimensions with appreciable rating variance, issue generality ($\kappa_\ell{=}0.40$, $\kappa_q{=}0.48$) and legal language ($\kappa_\ell{=}0.64$, $\kappa_q{=}0.74$), $\kappa$ behaves as expected and indicates moderate-to-strong agreement. On the remaining dimensions, however, $\kappa$ either drops close to zero or turns negative; in the case of summary form it reaches $\kappa_\ell{=}\kappa_q{=}-0.11$ despite an exact-agreement rate of $78.6\%$ and within-one-point agreement of $100\%$.

This pattern is a textbook instance of the so-called \emph{kappa paradox} \citep{feinstein1990high,wongpakaran2013comparison}. When one rating category accounts for the great majority of observations, the marginal-based estimate of chance agreement $p_e$ that enters $\kappa{=}(p_o-p_e)/(1-p_e)$ approaches the observed agreement $p_o$, and small fluctuations in either term swing $\kappa$ over a wide range, including into negative values. In our data, both annotators rate almost every issue 4 or 5 on the four summary dimensions and on the reasoning, outcome, and fact satisfaction scores, so $\kappa$ in this regime measures the rarity of disagreement rather than its substance. The same observation applies, less dramatically, to the Pearson statistics. Gwet's $\mathrm{AC2}$ \citep{gwet2014handbook} is designed precisely to be robust to this regime, and the values reported in Table~\ref{tab:quality_scores} of the main text give a more faithful picture of the level of agreement reached by the two annotators. For the citation-reason metric reported in Table~\ref{tab:cit_reason}, whose per-issue counts are well spread across the integer range $\{0,1,2,\ldots\}$ rather than saturated, the kappa paradox does not arise: $\kappa_\ell{=}0.91$, $\kappa_q{=}0.85$, and $r{=}0.86$, all consistent with the high agreement reported by the count-based statistics.

\begin{table}[!htbp]
\centering
\caption{Chance-corrected agreement statistics for the Likert dimensions ($N{=}20$, $28$~present issues). $\kappa_\ell$, $\kappa_q$: linearly and quadratically weighted Cohen's $\kappa$: two-way random-effects, absolute-agreement, single-measures intraclass correlation; $r$: Pearson correlation. In the case of outcome satisfaction, both annotators assigned $5$ to every issue, making variance-based statistics undefined.}
\label{tab:quality_scores_kappa}
\small
\begin{tabular}{l ccc}
\toprule
Dimension & $\kappa_\ell$ & $\kappa_q$ &  $r$ \\
\midrule
Issue generality        &  0.40 &  0.48 &   0.63 \\
Legal language          &  0.63 &  0.74 &    0.78 \\
Summary correctness     &  0.35 &  0.34 &    0.38 \\
Summary form            & $-$0.11 & $-$0.11 &  $-$0.11 \\
Summary completeness    &  0.18 &  0.08 &    0.08 \\
Summary satisfaction    &  0.18 &  0.08 &    0.09 \\
Fact satisfaction       & 0.45 & 0.40 & 0.46\\
Reasoning satisfaction  &  0.24 &  0.13 &  0.17 \\
Outcome satisfaction    & -- & -- & -- \\
\bottomrule
\end{tabular}
\end{table}

\section{Worked extraction example}
\label{app:worked_example}

This appendix reproduces in full the XML extraction summarised in Listing~\ref{lst:snail_farmer} in the main text, from judgment n.~318/2024 of the Corte di Giustizia Tributaria di primo grado di Teramo.

\begin{lstlisting}[caption={Full extraction for the worked example.}, label={lst:snail_farmer_full}]
<issues>
  <issue title="VAT deductibility for an agricultural business not yet productive" id="Q1">
    <text>Whether the purchase of a capital asset (agricultural tractor) by a newly established agricultural business, not yet productive, is connected to the entrepreneurial activity and therefore deductible pursuant to Articles 4 and 5 of Presidential Decree 633/72.</text>
    <issue_outcome in_favor_of="taxpayer">VAT is deductible because the purchase is related to the entrepreneurial activity, even in the absence of immediate taxable operations.</issue_outcome>
    <factual_premises>
      <item>The appellant began the snail-farming activity in 2020, an activity that requires time to become productive.</item>
      <item>In 2022, the appellant purchased an agricultural tractor for land preparation and for procuring plant-based feed necessary for the farming activity.</item>
      <item>The Revenue Agency denied the VAT refund, arguing that the activity had not yet started and that the tractor was not included in the depreciable assets list.</item>
      <item>The appellant provided documentation (company registry extract, invoices for the tractor and for snail purchase) to demonstrate the relevance of the purchase to the activity.</item>
    </factual_premises>
    <legal_references>
      <item id="D1" type="caselaw" ref="Cassazione n. 26689/2022; n. 7440/2021; n.15570/2023"/>
      <item id="D2" type="caselaw" ref="CJEU case C-110/94; case C-37/95; case C-400/98; case C-263/15"/>
      <item id="D3" type="caselaw" ref="CJEU case C-672/16"/>
      <item id="D4" type="leg" ref="Articles 4 and 5 DPR 633/72"/>
      <item id="D5" type="leg" ref="Article 2135 of the Civil Code"/>
    </legal_references>
    <citation_reason>
      <item ref_id="D1">National case law states that the right to deduct VAT is linked to the relevance of the asset to the entrepreneurial activity, even in the absence of immediate taxable operations, provided the asset is functional to the organisation of the business.</item>
      <item ref_id="D2">The EU Court of Justice reiterates that the right to deduct exists for investment expenses incurred in view of future entrepreneurial operations, without waiting for the actual start of the activity.</item>
      <item ref_id="D3">The EU Court of Justice emphasises that VAT deductibility is determined by the purchase of the asset, not by its immediate use, in order to ensure tax neutrality.</item>
    </citation_reason>
    <judge_reasoning>The judge considers that the purchase of the tractor is connected to the taxpayer's agricultural activity, despite the absence of immediate taxable operations, as it is functional to land preparation and the farming process. Both national and EU case law confirm that VAT deduction does not require the actual start of the activity, but only the relevance of the asset to entrepreneurial purposes. The Revenue Agency's objections, based on non-probative indicators, do not override this principle. Therefore, the denial is annulled and the VAT refund is granted.
    </judge_reasoning>
    <summary>The taxpayer, who operates a snail-farming business, purchased a tractor in 2022 to prepare the land, despite production not yet having begun. The Italian Revenue Agency denied the VAT refund, considering the activity not yet concretely started. The judge, referring to Article 30 of Presidential Decree 633/72 and case law (Civil Supreme Court no. 26689/2022 and EU Court of Justice C-110/94), held that deduction is admissible for assets necessary for the organisation of the business, even during the preparatory phase, provided there is an objective link with the planned activity. The ruling annulled the denial and recognised the deductibility.
    </summary>
  </issue>
</issues>
\end{lstlisting}
\section{Sensitivity analysis: run-to-run consistency}
\label{app:sensitivity}

As noted in Section~\ref{sec:LLM_extraction}, the extraction model is not
deterministic: even at zero temperature and with a fixed random seed, repeated
calls on identical inputs can yield different outputs, a phenomenon also
documented by \citet{blair2025llms}. All results in the main text are based on a
single run of the pipeline, and it is natural to ask how much they would change
under replication. In this appendix we quantify this run-to-run variability treating it in the same way as the inter-annotator disagreement of
Section~\ref{sec:results}, so that the two sources of variation can be compared.

\paragraph{Protocol.}
We ran the extraction pipeline a second time on the $50$ test judgments, under
identical settings (same prompts, model version, temperature, and seed),
obtaining a second collection of XML extractions. We denote by $R_1$ the run used
throughout the main validation and by $R_2$ the replication. A tax-law expert
then compared the two runs with the same set-based procedure adopted for the
human validation: for each judgment the issues of $R_1$ and $R_2$ were aligned by
content, and within each pair of aligned issues the corresponding legal
references were aligned, counting individual cited documents (a single
\verb|<item>| may bundle several, e.g.\ \emph{art.~21, commi 1 e 2}). Treating
$R_1$ and $R_2$ as two annotators, we report precision, recall,
and F1 with the definitions of Section~\ref{sec:validation_method}: for a system
run $S$ and a ground-truth run $G$, $\mathrm{P}_{S\mid G}=|S\cap G|/|S|$ and
$\mathrm{R}_{S\mid G}=|S\cap G|/|G|$, so the two directions exchange precision and
recall but share the same F1. Citations are evaluated within the issues on which
the two runs agree, the analogue of the \emph{present issues} of the main text.

Two features of this set-up should be kept in mind. First, the comparison is
carried out on the \emph{raw} LLM outputs, before the hallucination filter of
Section~\ref{sec:results_hallucination}, whereas the citation results in the main
text are computed \emph{after} filtering; since the filter removes references that
do not appear in the judgment, and different runs fabricate different references,
part of the citation variability reported below would be eliminated by the
filter. We chose to compare the raw outputs since we want to measure the LLM's sensitivity in isolation.  Second, the alignment is itself an imperfect, manual step: we have no
control over exactly which issues each run extracts, and the annotator must match
issues that are phrased differently and are not necessarily in a one-to-one correspondence. In
particular, the two runs do not extract identical issues, and even within an
aligned pair the issue formulation can differ slightly between runs, shifting the
set of references relevant to it. Part of the disagreement on references is
therefore inherited from the disagreement on issues rather than being a property
of citation extraction alone.

\paragraph{Extraction volume.}
The two runs are close in volume (Table~\ref{tab:sens_counts}), with the
replication $R_2$ slightly more prolific: it extracts $1.66$ issues per judgment
against $1.56$ for $R_1$ ($83$ vs $78$ issues in total), and $4.11$ vs $3.70$
legal references per aligned issue. The difference is small and does not, on its
own, indicate a systematic shift between runs.

\begin{table}[!htbp]
\centering
\caption{Average number of items per unit (totals in parentheses), per run.
The $63$ aligned issues are those identified by both runs.}
\label{tab:sens_counts}
\begin{tabular}{l cc}
\toprule
 & Issues per judgment & Refs per aligned issue \\
\midrule
$R_1$ (original)    & 1.56 (78) & 3.70 (233) \\
$R_2$ (replication) & 1.66 (83) & 4.11 (259) \\
\bottomrule
\end{tabular}
\end{table}

\paragraph{Issue extraction.}
The two runs identify the same legal issues with F1${=}78.3\%$
(Table~\ref{tab:sens_issues}). Of the $50$ judgments, $30$ ($60\%$) are segmented
identically; on the remaining $20$ at least one issue is found by only one of the
two runs. The two runs share $63$ issues.

\begin{table}[!htbp]
\centering
\caption{Run-to-run agreement on issue extraction ($50$ judgments). (ref.\ system $\mid$ ground truth). The two directions share the same F1.}
\label{tab:sens_issues}
\begin{tabular}{l ccc}
\toprule
Comparison & P & R & F1 \\
\midrule
$R_1 \mid R_2$ & 80.8 & 75.9 & 78.3 \\
$R_2 \mid R_1$ & 75.9 & 80.8 & 78.3 \\
\bottomrule
\end{tabular}
\end{table}

\paragraph{Citation extraction.}
Within the $63$ aligned issues, the two runs agree on legal references with
F1${=}73.2\%$ (Table~\ref{tab:sens_citations}). Agreement is far from complete:
$7$ of the $63$ aligned issues ($11\%$) share no reference at all.

\begin{table}[!htbp]
\centering
\caption{Run-to-run agreement on citation extraction, within the $63$ aligned
issues. References are counted as individual cited documents.}
\label{tab:sens_citations}
\begin{tabular}{l ccc}
\toprule
Comparison ($S \mid G$) & P & R & F1 \\
\midrule
$R_1 \mid R_2$ & 77.3 & 69.5 & 73.2 \\
$R_2 \mid R_1$ & 69.5 & 77.3 & 73.2 \\
\bottomrule
\end{tabular}
\end{table}

\paragraph{Comparison with inter-annotator agreement.}
The interest of these figures lies in the comparison with the human results of
Section~\ref{sec:results}, collected in Table~\ref{tab:sens_comparison}. Two runs
of the same model agree with each other \emph{less} than two human experts do:
on issue extraction the run-to-run F1 ($78.3\%$) sits below the inter-annotator
F1 ($88.1\%$), and on citations the gap is much wider ($73.2\%$ vs $97.1\%$). At
the same time, the run-to-run agreement is of the same order as the agreement
between the model and a human expert, falling within the LLM-vs-annotator range
on both issues ($72.7$--$83.6\%$) and citations ($73.8$--$74.8\%$). A substantial
part of the gap between the model and the human annotators is thus not a stable,
systematic difference but the model's own stochasticity; for citations in
particular, run-to-run variance is as large as the model--human gap itself.

\begin{table}[!htbp]
\centering
\caption{Agreement (F1, \%) under three regimes: between the two human annotators
(IAA), between the model and each annotator (LLM--Human), and between two runs of
the model (Run--Run). Human figures are reproduced from
Tables~\ref{tab:issue_pairwise} and~\ref{tab:cit_pairwise}.}
\label{tab:sens_comparison}
\begin{tabular}{l cc}
\toprule
Regime & Issue F1 & Citation F1 \\
\midrule
Human--Human (IAA)        & 88.1 & 97.1 \\
LLM--Human (range)        & 72.7--83.6 & 73.8--74.8 \\
Run--Run ($R_1$ vs $R_2$) & 78.3 & 73.2 \\
\bottomrule
\end{tabular}
\end{table}

These figures are conservative on two counts. The citation comparison is run on
raw outputs: the hallucination filter, which removes the fabricated references
that differ from run to run, would raise the measured agreement, so the
variability of the whole pipeline's output is smaller than
Table~\ref{tab:sens_citations} suggests. And part of the reference disagreement
is inherited from the imperfect alignment of issues across runs rather than from
citation extraction proper. The practical implication is, however, clear:
single-run scores are affected by the stochasticity of the model. Averaging over several judgments provides a way of reducing this noise.
\section{Rule-based citation extraction as a baseline}
\label{app:linkoln_baseline}

Section~\ref{sec:design_pressures} argues that a citation network built directly from judgments would include many references that are not central to the court's reasoning. Here we quantify that claim, and at the same time ask how much of the benefit of the LLM extraction could be obtained by a purely rule-based pipeline. We compare three ways of attaching a set of cited documents to a judgment: (i) all citations parsed by Linkoln from the full text; (ii) all citations parsed by Linkoln from the ``Motivi della decisione'' section, located with the same regex-based matcher used in preprocessing (Section~\ref{sec:dataset}); (iii) the citations extracted by our pipeline, after the hallucination filter. The reference standard is the set of citations that the experts judged relevant to the issues of the judgment.

\paragraph{Protocol.}
The comparison is restricted to judgments where (a) the annotators flagged no missing issues (otherwise their per-issue citation lists would not cover the entire judgment) and (b) the section heading is detectable. Of the $50$ validation judgments, $39$ satisfy (a), $39$ satisfy (b), and $31$ satisfy both ($19$ annotated by $A_1$, $20$ by $A_2$). Because the expert citations are recorded as free text, we parse them with Linkoln as well, and match citations automatically at the document level (number and year, completed with URN identifiers where needed). This automatic matching differs from the manual matching underlying Table~\ref{tab:cit_pairwise}, so absolute scores are not comparable between the two analyses; within this appendix, however, all three tiers are evaluated identically. References that the parser cannot represent--- general legal principles, administrative practice, and generic collective references (e.g.\ \emph{giurisprudenza consolidata})---are excluded from all tiers.

\begin{table}[!htbp]
\centering
\caption{Citations per judgment and pooled precision/recall of three extraction tiers against the expert-relevant citations, on the $31$ qualifying judgments. Automatic document-level matching; principles excluded. In brackets: $95\%$ judgment-level bootstrap confidence intervals.}
\label{tab:linkoln_baseline}
\small
\begin{tabular}{l ccc}
\toprule
 & Cit./judgment & P & R \\
\midrule
\multicolumn{4}{l}{\emph{vs $A_1$ ($N{=}19$; $4.05$ relevant citations/judgment)}}\\
Full judgment (Linkoln)      & 6.74 & 57.8 [43.8, 69.3] & 96.1 [88.9, 100.0] \\
Motivi section (Linkoln)     & 4.79 & 81.3 [70.4, 90.7] & 96.1 [88.9, 100.0] \\
LLM extraction (post-filter) & 3.53 & 88.1 [73.8, 97.9] & 76.6 [68.1, 90.3] \\
\midrule
\multicolumn{4}{l}{\emph{vs $A_2$ ($N{=}20$; $3.40$ relevant citations/judgment)}}\\
Full judgment (Linkoln)      & 7.60 & 39.5 [26.9, 53.5] & 88.2 [77.8, 98.4] \\
Motivi section (Linkoln)     & 5.60 & 53.6 [33.8, 80.0] & 88.2 [77.8, 98.4] \\
LLM extraction (post-filter) & 3.55 & 74.6 [54.0, 91.0] & 77.9 [66.7, 89.7] \\
\bottomrule
\end{tabular}
\end{table}

\paragraph{Results.}
Table~\ref{tab:linkoln_baseline} shows a consistent gradient. The full text contains the most citations per judgment and the lowest precision, confirming the dilution argument. Restricting the parser to the reasoning section is a strong and essentially free improvement: it removes the references contained in the parties' submissions at no cost in recall (every relevant citation found in the full text already lies in the section). The LLM extraction is the most selective tier and the most precise, and gives up $10$--$20$ points of recall in exchange. We believe the recall gap can be bridged by using more powerful LLMs.

On the subset where the section heading can be located, a Linkoln-only pipeline is a reasonable low-cost alternative for \emph{judgment-level} citation networks: it requires no LLM calls and achieves higher recall than our pipeline. The case for the LLM extraction rests on what no section-level method provides: (i) the attribution of each citation to the specific issue it serves, which is what makes the edges of the network interpretable and is a prerequisite for the issue-level uses of Section~\ref{sec:design_pressures}; (ii) the presence of additional \verb|<citation_reason>| elements attached to each citation; and (iii) applicability to the whole corpus, whereas the section heading is detected in only $39$ of the $50$ validation judgments ($78\%$). These results are based on $31$ judgments and automatic matching, and should be read as indicative.
\section{Model selection}
\label{app:model_selection}

The choice of extraction model was made through an informal, qualitative
comparison carried out during the development phase, prior to and separately from
the validation of Section~\ref{sec:validation_method}. Because the pipeline must
run over several hundred thousand judgments, the goal was not to identify the
single most capable model but a sufficiently cheap model whose output the legal
experts judged acceptable. The assessment was deliberately lightweight: we did not
compute the set-based metrics of the main text, which would have required a second
full annotation effort, but relied on the two tax-law experts reading and
commenting on the extractions produced by each candidate on a small development set
of 10 decisions, disjoint from the $50$ validation judgments.

We compared a range of open-weight, cost-efficient models against a top-range
proprietary reference (Claude Sonnet~4.5). The open-weight candidates included
several DeepSeek~V3 variants (V3-0324, V3.1, and V3.2-exp), Llama~3.3~70B,
Gemma~3~27B, Qwen QwQ~32B, Mistral Devstral, and Kimi-K2. Models were accessed
through third-party API providers.

For each candidate the experts considered, informally: (i) the quality of the
segmentation of a judgment into issues, penalising both artificial splitting and
the omission of genuine issues; (ii) the precision and appropriateness of the legal
language in the free-text fields; (iii) the faithfulness of the extracted
\verb|<legal_references>|, i.e. the tendency to fabricate or misattribute
citations (e.g. misattributing the references relevant for one issue to another of the same judgment); and (iv) price.

Among the models reaching a quality acceptable to the experts on criteria
(i)--(iv), DeepSeek~V3-0324 offered the best trade-off with cost and was selected
for the bulk of the extraction. Claude Sonnet~4.5 produced better output but at roughly twenty times the price (Section~\ref{sec:LLM_extraction}), which is not sustainable at corpus scale. The
smaller open-weight models (Llama~3.3~70B, Gemma~3~27B, QwQ~32B, Devstral) were
less reliable on issue segmentation and on the precision of legal language, while
the remaining large open-weight models did not offer a clear advantage over
DeepSeek at comparable cost. The slightly newer DeepSeek~V3.1 was adopted only for
the lightweight title-correction step (Section~\ref{sec:LLM_extraction}). We stress
that this comparison is qualitative and was intended to support an engineering
decision under a cost constraint, not to establish a ranking of the models on the
extraction task.
\end{document}